\documentclass[10pt,twocolumn,letterpaper]{article}

\usepackage{iccv}
\usepackage{times}
\usepackage{epsfig}
\usepackage{graphicx}
\usepackage{amsmath}
\usepackage{amssymb}
\usepackage[utf8]{inputenc} 
\usepackage[T1]{fontenc}    
\usepackage{url}            
\usepackage[pagebackref=true,breaklinks=true,letterpaper=true,colorlinks,bookmarks=false]{hyperref}
\usepackage{booktabs}       
\usepackage{amsfonts}       
\usepackage{amsmath}        
\usepackage{mathtools}
\usepackage{nicefrac}
\usepackage{algorithm}
\usepackage{algpseudocode}
\usepackage{microtype}      

\usepackage{tabularx} 

\usepackage[capitalize]{cleveref}

\usepackage{mlmacros}       

\usepackage{threeparttable}
\usepackage{enumitem}

\usepackage{siunitx}
\usepackage{subfig}
\usepackage{wrapfig}
\usepackage{dblfloatfix}
\usepackage{soul}
\usepackage[normalem]{ulem} 

\usepackage{tikz}
\usetikzlibrary{shapes,arrows,positioning,matrix,snakes}
\usetikzlibrary{decorations.pathmorphing}
\tikzset{snake it/.style={decorate, decoration=snake}}

\usepackage[toc,page]{appendix}
\crefname{supplementary}{Appendix}{Appendices}

\DeclareMathOperator*{\argmax}{\arg\!\max} %
\probdists{p,q}
\MkProbDist{Cat}{\mathrm{Cat}}

\newcommand{\PLH}{{\mkern-2mu\times\mkern-2mu}}

\iccvfinalcopy 

\def\iccvPaperID{4372} 

\ificcvfinal\fi

\title{Calibrated Adversarial Refinement for Stochastic Semantic Segmentation}

\begin{document}

\author{%
  Elias Kassapis$^{1,2,*}$ \quad Georgi Dikov$^{2}$ \quad Deepak K. Gupta$^{1}$ \quad Cedric Nugteren$^{2}$ \\ \\
  $^{1}$ Informatics Institute, University of Amsterdam, The Netherlands \\
  $^{2}$ TomTom, Amsterdam, The Netherlands \\
  $^*$ \texttt{kassapiselias@hotmail.co.uk} \\
}


\maketitle
\ificcvfinal\thispagestyle{empty}\fi


\begin{abstract}

In semantic segmentation tasks, input images can often have more than one plausible interpretation, thus allowing for multiple valid labels. To capture such ambiguities, recent work has explored the use of probabilistic networks that can learn a distribution over predictions. However, these do not necessarily represent the empirical distribution accurately. In this work, we present a strategy for learning a calibrated predictive distribution over semantic maps, where the probability associated with each prediction reflects its ground truth correctness likelihood. To this end, we propose a novel two-stage, cascaded approach for calibrated adversarial refinement: (i) a standard segmentation network is trained with categorical cross entropy to predict a pixelwise probability distribution over semantic classes and (ii) an adversarially trained stochastic network is used to model the inter-pixel correlations to refine the output of the first network into coherent samples. Importantly, to calibrate the refinement network and prevent mode collapse, the expectation of the samples in the second stage is matched to the probabilities predicted in the first. We demonstrate the versatility and robustness of the approach by achieving state-of-the-art results on the multigrader LIDC dataset and on a modified Cityscapes dataset with injected ambiguities. In addition, we show that the core design can be adapted to other tasks requiring learning a calibrated predictive distribution by experimenting on a toy regression dataset. We provide an open source implementation of our method at \url{https://github.com/EliasKassapis/CARSSS}.

\end{abstract}

\section{Introduction}

Real-world datasets are often riddled with ambiguities, allowing for multiple valid solutions for a given input. These can emanate from an array of sources, such as sensor noise, occlusions, inconsistencies during manual data annotation, or an ambiguous label space~\cite{lee2016stochastic}.
Despite the fact that the empirical distribution can be multimodal, the majority of the research encompassing semantic segmentation focuses on optimising models that assign only a single solution to each input image~\cite{unet,tiramisu,gated-shape-cnn,chen2017deeplab,chen2017rethinking,CY2016Attention,CB2016Semantic,CP2015Semantic}, and are thus often incapable of capturing the entire empirical distribution.

\begin{figure}[t]
  \centering
  \input{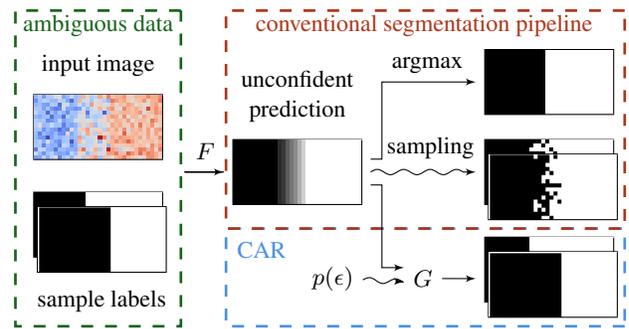}
  \caption{Conceptual diagram of stochastic semantic segmentation: blue and red pixels are separable by several different vertical boundaries, resulting in multiple valid labels. $F$ is a network parametrising a factorised categorical likelihood, which captures the pixelwise data ambiguity. Extracting the mode via the $\argmax$ operation deterministically yields a single coherent prediction, while direct sampling gives multiple incoherent ones. Instead, calibrated adversarial refinement (CAR) uses a second stochastic adversarial network, $G$, which refines the output of $F$ into diverse, coherent labels.}
  \label{fig:concept}
\end{figure}

These approaches typically model each pixel independently with a factorised categorical likelihood, and therefore do not consider inter-pixel correlations during sampling (see~\cref{fig:cs_entropy}b in~\cref{sup:more_cs}). Further, since maximising the likelihood on noisy datasets leads to unconfident predictions in regions of label inconsistencies, direct sampling yields incoherent semantic maps. 
Alternatively, coherent predictions can be obtained by applying the $\argmax$ function, essentially extracting the mode of the likelihood. 
This, however, comes at the cost of limiting the model's representation capabilities to deterministic, one-to-one mappings between inputs and outputs (see~\cref{fig:concept}).

Here we consider the problem of \textit{stochastic semantic segmentation}: the task of semantically segmenting ambiguous images with an arbitrary number of
valid labels, each with a distinct probability of occurrence.
To this end, an ideal model should capture joint pixel dependencies and leverage uncertainty information to sample multiple coherent hypotheses. Further, it is important that the empirical occurrence frequency of each sampled segmentation variant reflects its ground truth correctness likelihood; that is, the predictive distribution should be calibrated~\cite{calibration,kull2019beyond}. Such a system would be especially useful for semi-automatic safety-critical applications, \eg medical imaging and map making, where it is crucial to identify ambiguous input and cross-examine all possible interpretations and their corresponding likelihoods before making an important decision~\cite{data:lidc, menze2014multimodal,litjens2012pattern}.

In this work, we introduce calibrated adversarial refinement (CAR): a two-stage, cascaded framework for learning a calibrated, multimodal predictive distribution.
In the first stage, we train a standard network with categorical cross entropy to estimate pixelwise class probabilities, as well as the associated aleatoric uncertainty estimates~\cite{kendall2017uncertainties}. In the second, an adversarial network, capable of modelling the inter-pixel dependencies, is used to sample realistic, coherent predictions (see bottom of~\cref{fig:concept}). The sample diversity is then calibrated relatively to the distribution predicted in the first stage, via an additional loss term.~Our~key~contributions~are:

\begin{itemize}[noitemsep]
    \item We propose a novel cascaded architecture for adversarial refinement that allows sampling of an arbitrary number of coherent segmentation maps.
    \item We introduce a novel loss term, called the \textit{calibration loss}, that facilitates learning of calibrated stochastic mappings and mitigates mode collapse in conditional adversarial learning.
    \item Our model can be trained independently or used to augment any black-box semantic segmentation model.
\end{itemize}

\section{Related work}

Straightforward strategies towards learning multiple predictions include ensembling~\cite{lakshminarayanan2017simple,kamnitsas2017ensembles} or using multiple prediction heads~\cite{rupprecht2017learning}. Even though these approaches can capture a diverse set of sampled predictions, they are limited to only a fixed number of samples. 
Alternatively, a probability distribution over the outputs can be induced by activating dropout during test time~\cite{gal2016dropout}. 
This method does offer useful uncertainty estimates over the pixel-space~\cite{mukhoti2018evaluating}, however, it has been demonstrated~\cite{pix2pix, rel:probUnet} that it introduces only minor stochasticity in the output and returns incoherent samples.

Bhattacharyya \etal (2018)~\cite{bhattacharyya2018bayesian} identify the maximum likelihood learning objective as the cause for this issue in dropout Bayesian neural networks~\cite{gal2016bayesian}. They postulate that under cross entropy optimisation, all sampled models are forced to explain the entirety of the data, thereby converging to the mean solution. 
They propose to mitigate this issue using variational inference and replacing cross entropy with an adversarial loss term parametrising a synthetic likelihood~\cite{rosca2017variational}. This renders the objective function conducive to multimodality but, unlike our method, requires the specification a weight prior and variational distribution family. 

Kohl \etal (2018)~\cite{rel:probUnet} take an orthogonal approach in combining a U-Net~\cite{unet} with a conditional variational autoencoder (cVAE)~\cite{vae} to learn a distribution over semantic labels. Hu \etal (2019)~\cite{multipleannotations} build on~\cite{rel:probUnet} by leveraging intergrader variability as additional supervision. Even though this improves performance, a major limitation of this approach is the requirement of a priori knowledge of all the modes in the data distribution, often unavailable in real-world datasets.
Alternatively, subsequent work in \cite{kohl2019hierarchical} and~\cite{baumgartner2019phiseg} improve the diversity of the samples of~\cite{rel:probUnet} by modelling the data on several scales of the image resolution. In more recent work, Monteiro \etal (2020)~\cite{monteiro2020stochastic} take a different path, proposing a single network to parametrise a low-rank multivariate Gaussian distribution which models the inter-pixel and class dependencies in the logit space. This method does improve efficiency during inference, however, a low-rank parametrisation imposes a constraint on the sample complexity. In contrast to all these methods, ours uses an adversarial loss term which has been shown to elicit superior structural qualities than cross entropy~\cite{luc2016semantic, elgan, gambler}.

In the more general domain of image-to-image translation, 
hybrid models have been proposed using adversarially trained cVAEs~\cite{multimodal_im-2-im,cvae-gan} to learn a distribution over a latent code that encodes multimodality, allowing sampling of diverse yet coherent predictions. A common hurdle in such conditional generative adversarial networks (cGANs)  
is that simply incorporating a noise vector as an additional input often leads to mode collapse. This occurs due to the lack of regularisation between noise input and generator output, allowing the generator to learn to ignore the noise vector~\cite{pix2pix}. This is commonly resolved by using supplementary cycle-consistency losses~\cite{unsupervised_multimodal,diverse_image2image,multimodal_im-2-im,cvae-gan}, as proposed by Zhu \etal (2017)~\cite{zhu2017unpaired}, or with alternative regularisation losses on the generator~\cite{yang2018diversity}. 
Nonetheless, these do not address the challenge of calibrating the predictive distribution.

\section{Method}
\label{sec:methods}

\subsection{Motivation}
\label{sec:background}

Semantic segmentation refers to the task of predicting a pixelwise class label $y \in \{1,\dots,K\}^{H \times W}$ given an input image $x \in \mathbb{R}^{H \times W \times C}$. For a dataset of $N$ image and label pairs, $\data = \{x_i, y_i\}_{i = 1}^N$, the empirical distribution $\p[\data]{y}{x}$ can be explicitly modelled through a likelihood $\q[\theta]{y}{x}$, parametrised by a softmax-activated convolutional neural network $F$ with weights $\theta$~\cite{unet, tiramisu}. One simple, yet effective way to learn the class probabilities is to express $y \in \{0, 1\}^{H \times W \times K}$ in a one-hot encoded representation and set $\q[\theta]$ as a pixelwise factorised categorical distribution:
\begin{equation}
    \q[\theta]{y}{x} = \prod_i^H \prod_j^W \prod_k^K F_\theta(x)_{i,j,k}^{y_{i,j,k}}.
\label{eq:likelihood}
\end{equation}
The parameters $\theta$ are then optimised by minimising the cross entropy between $\p[\data]$ and $\q[\theta]$, defined as:
\begin{equation}
    \mathcal{L}_{\mathrm{ce}}(\data, \theta) = -\expc[\p[\data]{x,y}]{\log \q[\theta]{y}{x}}.
\label{eq:loss_ce}
\end{equation}

When trained with~\cref{eq:loss_ce}, $F_\theta$ learns an approximation of \expc[\p[\data]]{y}{x}~\cite{bishop2006pattern}, thereby capturing the per-pixel class probabilities over the label that corresponds to a given input. At this point, the aleatoric uncertainty can be obtained by computing the entropy of the output of $F_\theta$, $\entropy{F_\theta(x)}$~\cite{kendall2017uncertainties}. 

As discussed in the introduction section and exemplified in~\cref{fig:concept}, neither sampling from the likelihood in~\cref{eq:likelihood}, nor extracting its mode are adequate solutions for stochastic semantic segmentation, where multiple valid predictions are sought. 
This issue can be partially addressed by adapting the framework of generative adversarial networks (GANs)~\cite{goodfellow2014generative} to the context of semantic segmentation, as proposed by~\cite{luc2016semantic}. Formally, this involves training a binary discriminator network $D$ to optimally distinguish between ground truth and predictions, while concurrently training a conditional generative network $G$ to maximise the probability that prediction samples $G(x)$ are perceived as real by $D$. Importantly, in contrast to explicit pixelwise likelihood maximisation, the adversarial setup learns an implicit sampler through $G$, capable of modelling the joint pixel configuration of the synthesised labels, and capturing both local and global consistencies present in the ground truth~\cite{luc2016semantic}. 

In practice, the generator loss is often complemented with the pixelwise loss from \cref{eq:loss_ce} to improve training stability and prediction quality~\cite{luc2016semantic,elgan,gambler}. However, we argue that the two objective functions are not well aligned in the presence of noisy data. While categorical cross entropy optimises for a single, mode averaging solution for each input $x$, thus encouraging high entropy in $\q[\theta]{y}{x}$ within noisy regions of the data, the adversarial term optimises for low-entropy, label-like output, and allows multiple solutions.
Therefore combining these losses in an additive manner, and enforcing them on the same set of parameters can be suboptimal---this prompts the generator to collapse to a deterministic output, as we show in~\cref{sec:evaluation} experimentally by using $\mathcal{L}_\mathrm{ce}$-regularised baselines. 

\subsection{Calibrated adversarial refinement}
\label{sec:car_method}

In this work, we propose to avert potential conflict between the cross entropy and adversarial losses by decoupling them in a two-stage, cascaded architecture. This consists of a \textit{calibration network} $F_\theta$, optimised with $\mathcal{L}_\mathrm{ce}$ from~\cref{eq:loss_ce}, the output of which is fed to a \textit{refinement network} $G_\phi$, optimised with an adversarial loss, which is in turn parametrised by an auxiliary discriminator $D_\psi$ trained with a binary cross entropy loss.

To account for the multimodality in the labels, we condition the refinement network on an additional extraneous noise variable $\epsilon \sim \gauss{0, 1}$, as done in the original GAN framework proposed by Goodfellow \etal~\cite{goodfellow2014generative}. In practice, we also condition the refinement network and the discriminator on the inputs $x$, however, we do not show this explicitly for notational convenience. 
More formally, using the non-saturated version of the adversarial loss~\cite{goodfellow2014generative}, the objectives for the refinement and discriminator networks are given by:
\begin{align}
    \begin{split}
    \mathcal{L}_{\mathrm{adv}}(\data, \theta, \phi) &= -\expc[\p[\data],\p[\epsilon]]{\log D_\psi(G_\phi(F_\theta(x), \epsilon))},
    \end{split}
    \\
    \begin{split}
    \mathcal{L}_{\mathrm{D}}(\data, \theta, \phi, \psi) &=
    -\expc[\p[\data],\p[\epsilon]]{\log\left(1 - D_\psi(G_\phi(F_\theta(x), \epsilon))\right)} \\ 
    &\qquad-\expc[\p[\data]]{\log D_\psi(y)}.
    \end{split}
\label{eq:loss_adv_new}
\end{align}

\begin{figure*}[t!]
    \centering
    \resizebox{0.99\textwidth}{!}{%
        \input{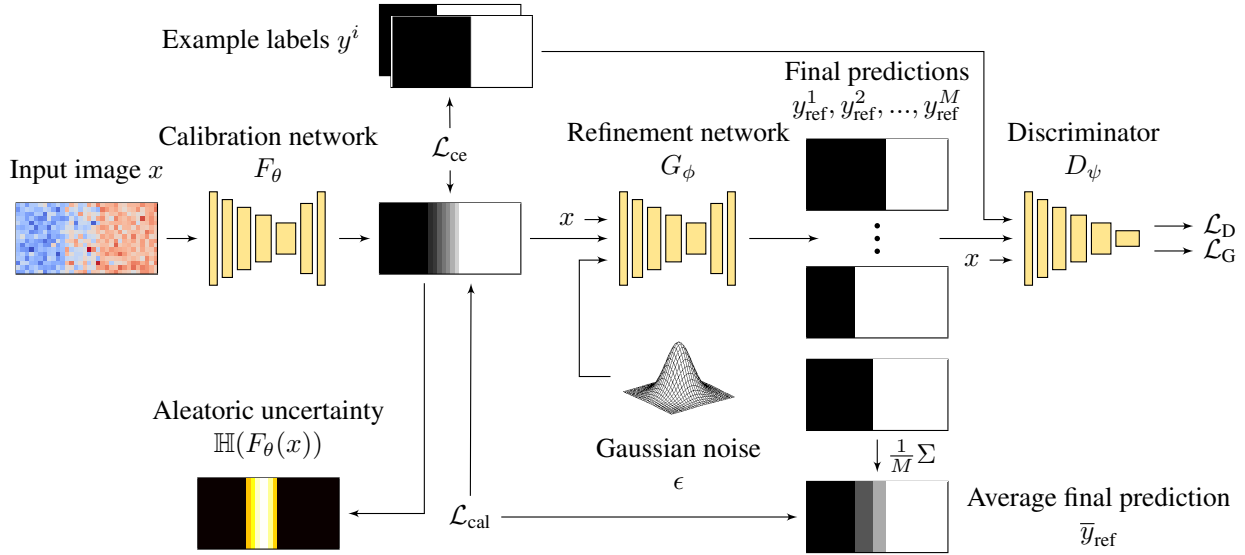}
    }
    \caption{CAR model diagram based on the example in~\cref{fig:concept}. First, the calibration network maps the input to a pixelwise distribution over the labels. This is then fed into the refinement network which samples an arbitrary number of diverse, crisp label proposals $y^1_\text{ref},\ldots, y^M_\text{ref}$. To ensure calibration, the average of the final predictions is matched with the calibration target from the first stage through the $\mathcal{L}_\text{cal}$ loss. Additionally, the aleatoric uncertainty can be readily extracted from the calibration target, \eg by computing the entropy $\entropy{F_\theta(x)}$.}
    \label{fig:cass_model}
\end{figure*}

To calibrate the predictive distribution, we impose diversity regularisation on $G_\phi$ by introducing a novel loss term, which we call the \textit{calibration loss}, that encourages the sample average $\overline{G}_{\phi}(F_\theta(x)) \coloneqq \expc[\p{\epsilon}]{G_\phi(F_\theta(x), \epsilon)}$ to match the pixelwise class probabilities predicted by $F_\theta(x)$. Here, $\overline{G}_{\phi}(F_\theta(x))$ serves as a factorised approximation to the implicit predictive distribution of the refinement network. To this end, we define an auxiliary categorical likelihood $\q[\phi]$ as:
\begin{equation}
    \q[\phi]{y}{F_\theta(x)} = \prod_i^H \prod_j^W \prod_k^K \overline{G}_{\phi}(F_\theta(x))_{i,j,k}^{y_{i,j,k}},
\end{equation}
and optimise $\phi$ using the proposed calibration loss, formulated as:
\begin{equation}
    \mathcal{L}_{\mathrm{cal}}(\data, \theta, \phi) = \expc[\p[\data],\q[\phi]]{\log \frac{\q[\phi]{y}{F_\theta(x)}}{\q[\theta]{y}{x}}}.
\label{eq:loss_cal}
\end{equation}
This loss term expresses the the Kullback-Leibler divergence, $\kl{\q[\phi]}{\q[\theta]}$\footnote{The choice of divergence is heuristically motivated and can be changed to fit different use-case requirements.}. Since both $\q[\phi]$ and $\q[\theta]$ are categorical distributions, the divergence can be computed exactly.

Notice that $\mathcal{L}_{\mathrm{cal}}$ optimises through an approximation of the expectation $\overline{G}_{\phi}(F_\theta(x))$, rather than a single sampled prediction, therefore the model is not restricted to learning a mode-averaging solution for each input $x$. Consequently, $\mathcal{L}_{\mathrm{cal}}$ is more compatible with $\mathcal{L}_{\mathrm{adv}}$ than $\mathcal{L}_{\mathrm{ce}}$ when a multimodal predictive distribution is desired.
The total loss for the refinement network then becomes:
\begin{equation}
    \mathcal{L}_{\mathrm{G}}(\data, \theta, \phi) = \mathcal{L}_{\mathrm{adv}}(\data, \theta, \phi) + \lambda \mathcal{L}_{\mathrm{cal}}(\data, \theta, \phi),
\label{eq:loss_total}
\end{equation}
where $\lambda \geq 0$ is a hyperparameter. \cref{fig:cass_model} shows the interplay of $F_\theta$, $G_\phi$ and $D_\psi$ and the corresponding loss terms.

Intuitively, the calibration network $F_\theta$ serves three main purposes: (i) it sets a calibration target used by $\mathcal{L}_\text{cal}$ to regularise the predictive distribution of $G_\phi$, (ii) it provides $G_\phi$ with an augmented representation of $x$ enclosing probabilistic information about $y$, (iii) it accommodates the extraction of sample-free aleatoric uncertainty maps.
The refinement network can therefore be interpreted as a stochastic sampler, modelling the inter-pixel dependencies to draw realistic samples from the explicit likelihood provided by the calibration network. Thus both the pixelwise class probability and object coherency are preserved.
This approach leads to improved mode coverage, training stability and increased convergence speed, as demonstrated in~\cref{sec:evaluation}.

\subsection{Practical considerations}

The gradients generated from the refinement network's loss function $\mathcal{L}_\mathrm{G}$ are prevented from flowing into the calibration network, to ensure that $F_\theta$ learns an unbiased estimate of $\expc[\p[\data]]{y}{x}$.
As a consequence the weights of $F_\theta$ can be kept fixed, while the adversarial pair $G_\phi$ and $D_\psi$ is being trained. This allows $F_\theta$ to be pretrained in isolation, thereby lowering the overall peak computational and memory requirements and improving training stability (see~\cref{alg:training,alg:inference} in ~\cref{sup:training} for an outline of the training and inference procedures). 
Further, computing $\mathcal{L}_\mathrm{cal}$ requires a Monte Carlo estimation of $\overline{G}_{\phi}(F_\theta(x))$, where the quality of the loss feedback increases with the sample count. 
This introduces a trade-off between training speed and prediction quality, however, modern deep learning frameworks allow for the samples to be subsumed in the batch dimension, and can therefore be efficiently computed on GPUs. 
Finally, our method can augment any existing black-box model $B$ for semantic segmentation, furnishing it with a calibrated multimodal predictive distribution. This can be done by conditioning $F_\theta$ on the output of $B$,~as~we~demonstrate~in~\cref{sec:evaluation_cs}.

\section{Experiments}
\label{sec:evaluation}

\begin{figure*}[t!]
    \centering
    \subfloat[]{{\includegraphics[width=0.32\textwidth]{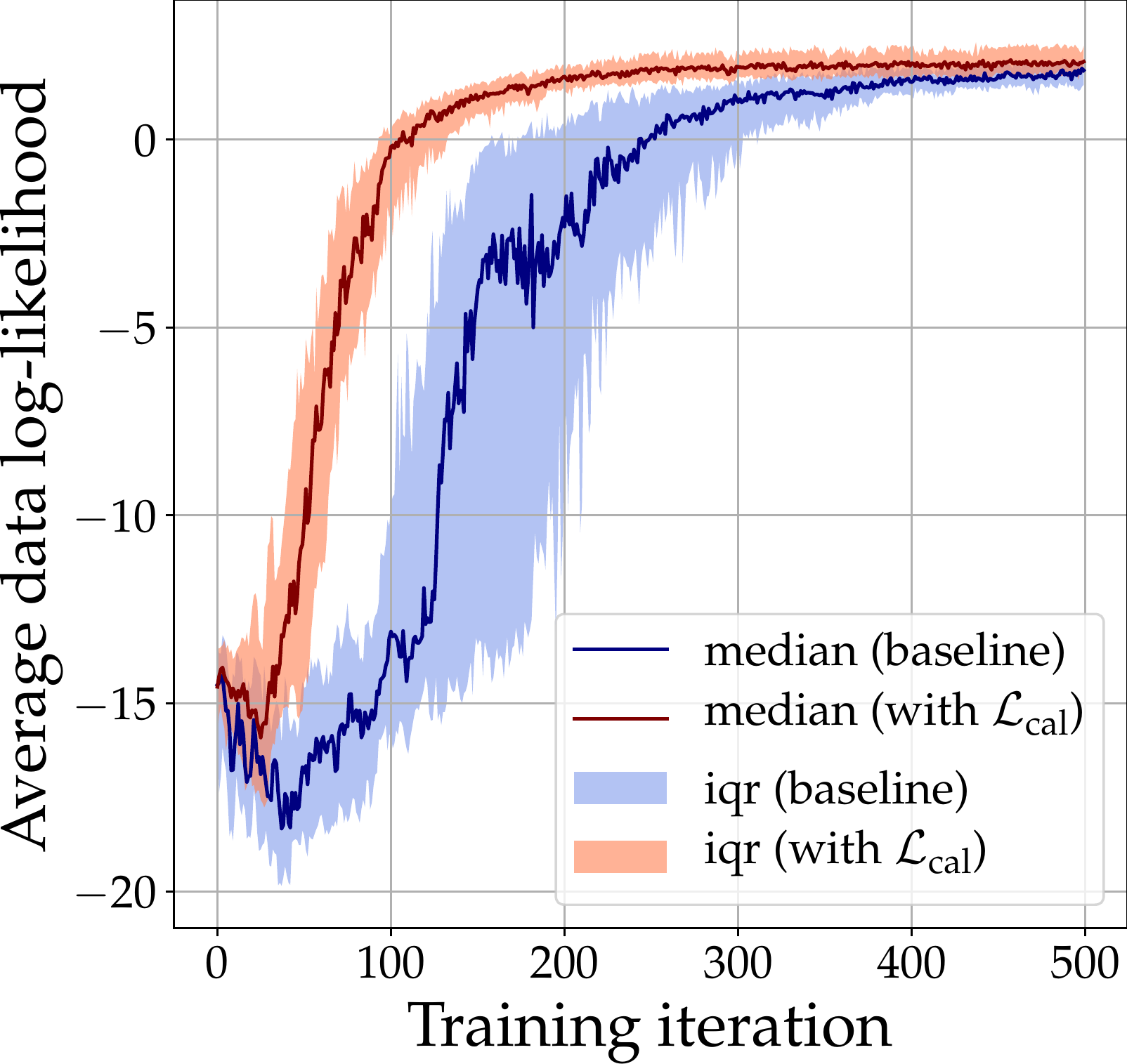} }}%
    \,
    \subfloat[]{{\includegraphics[width=0.32\textwidth]{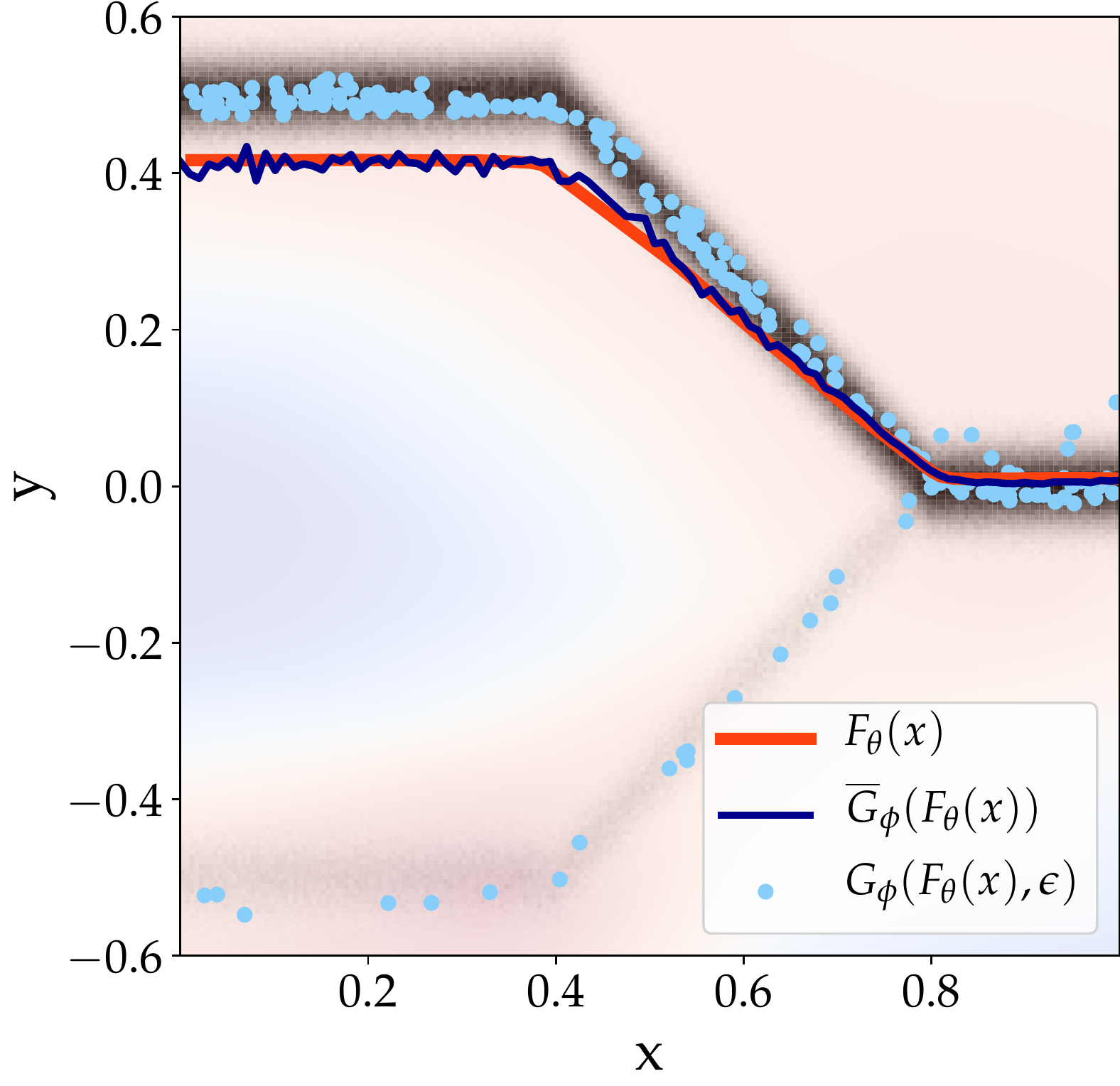} }}%
    \,
    \subfloat[]{{\includegraphics[width=0.32\textwidth]{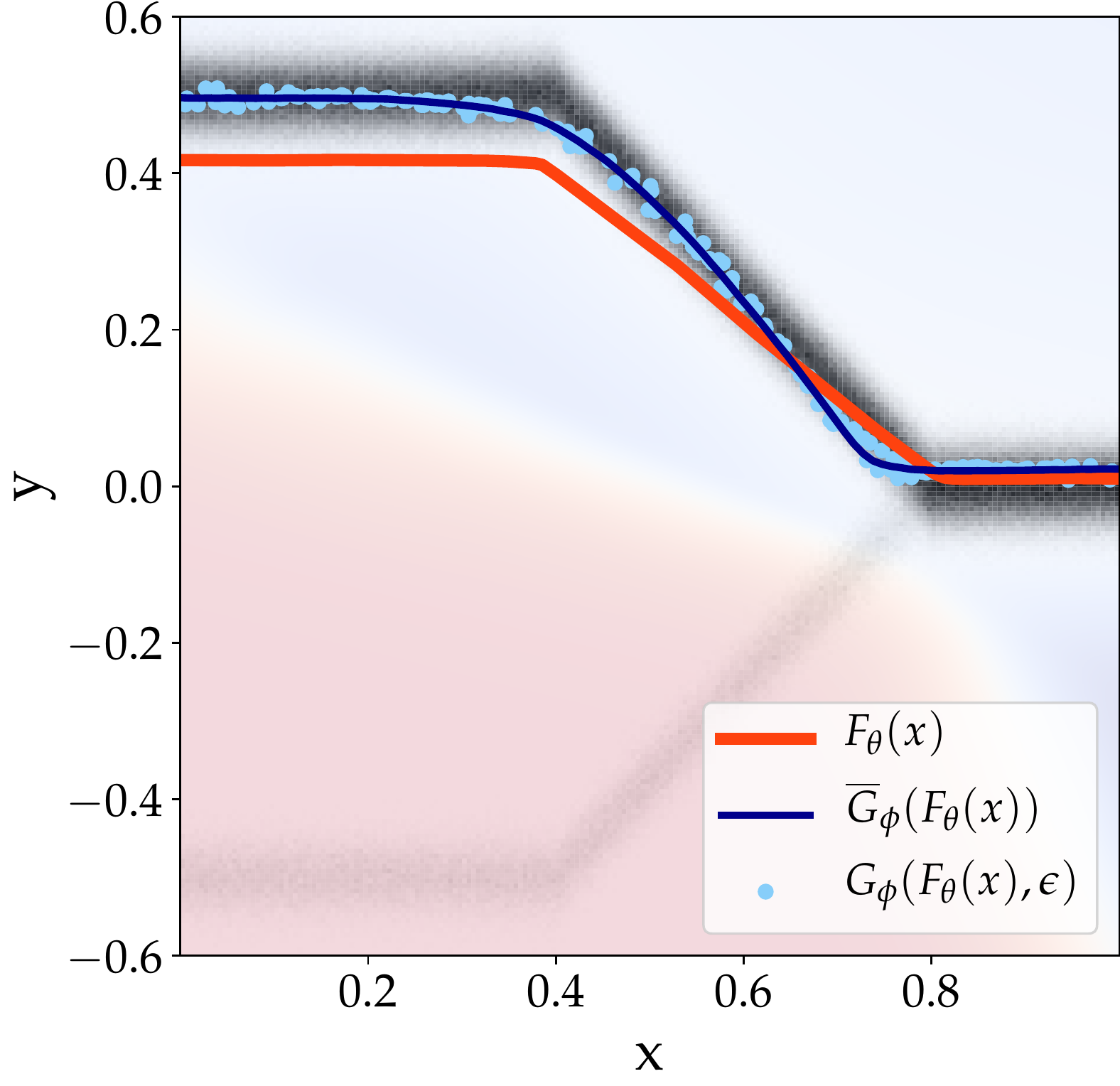} }}%
    \caption{\textbf{(a)} Median and interquartile range (iqr) over the data log-likelihood, averaged over all $9 \PLH 5 \PLH 2$ experiments. \textbf{(b)} High bias and noise configuration ($\pi=0.9,\ \sigma=0.03$) with calibration loss. The ground truth target is shown as black dots and the predicted samples as light blue dots. The predictions average in dark blue matches the calibration target in red. The discriminator output is shown in the background in shades of red (real) and blue (fake). \textbf{(c)} The same experiment configuration but without the proposed calibration loss, resulting in a mode collapse.}%
    \label{fig:toy_data}%
\end{figure*} 

\subsection{1D bimodal regression}
\label{sec:toy_regression}
We give intuitive insight into the mechanics of the proposed calibration loss by designing and experimenting on a simple one-dimensional regression task. To create the dataset, an input $x \in [0, 1]$ is mapped to $y \in \mathbb{R}$ as follows:
\begin{equation}
    y =  
    \begin{cases}
      0.5 - b + \epsilon, & x \in [0, 0.4) \\
      (-1)^b(-1.25x + 1) + \epsilon, & x \in [0.4, 0.8) \\
      \epsilon, & x \in [0.8, 1]
    \end{cases}
\end{equation}
where $b \sim \mathrm{Bernoulli}(\pi)$ and $\epsilon \sim \gauss{0, \sigma}$. We generate 9 different scenarios by varying the degree of mode selection probability $\pi \in \{0.5, 0.6, 0.9\}$ and the mode noise $\sigma \in \{0.01, 0.02, 0.03\}$.

For every data configuration, we use a 4-layer MLP for each of $F_{\theta}$, $G_{\phi}$ and $D_{\psi}$, and train with and without calibration loss by setting the coefficient $\lambda$ in~\cref{eq:loss_total} to 1 and 0, respectively.  Note that unlike the categorical likelihood used in semantic segmentation tasks, we employ a Gaussian likelihood with fixed scale of 1. This changes the formulation of both \cref{eq:loss_ce,eq:loss_cal} to mean squared error losses between ground truth labels $y$ and predictions $\hat{y}$ for $\mathcal{L}_{\mathrm{ce}}$, and between the output of the calibration network $F_{\theta}(x)$ and the average of multiple predictions $\overline{G}_{\phi}(F_\theta(x))$ for $\mathcal{L}_{\mathrm{cal}}$ (see~\cref{sup:regression_details}). Finally, all runs are trained with a learning rate of \num{1e-4} and each experiment is repeated five times.

The results, depicted in~\cref{fig:toy_data}, illustrate that when using calibration loss, the optimisation process shows improved training stability, converges faster, and results in better calibrated predictions in comparison to the non-regularised baseline. Notice that $\mathcal{L}_{\mathrm{cal}}$ also serves as an effective mechanism against mode collapse.
This effect is more pronounced in data configurations with higher bias. More plots of the individual experiments are shown in~\cref{sup:more_regression}.

\subsection{Stochastic semantic segmentation}
In this section we examine the capacity of our model to learn shape and class multimodality in real-world segmentation datasets. We begin by sharing essential implementation details below.

\paragraph{Network architectures} 
For the calibration network $F_\theta$, we use the SegNet~\cite{segnet} encoder-decoder architecture. For $G_\phi$, we designed a U-Net-style~\cite{unet} architecture with 4 down- and upsampling blocks, each consisting of a convolutional layer, followed by a batch normalisation layer~\cite{ioffe2015batch}, a leaky ReLU activation, and a dropout layer~\cite{srivastava2014dropout} with 0.1 dropout probability. We use a base number of 32 channels, doubled or halved at every down- and upsampling transition. To propagate the sampled noise vector $\epsilon$ to the output, we inject it into every upsampling block of the network in a linear manner. 
To do so, we project $\epsilon$ using two fully connected layers into scale and residual matrices with the same number of channels as the feature maps at the points of injection, and use these matrices to adjust the channel-wise mean and variance of the activations. This is similar to the mechanism used for adaptive instance normalisation~\cite{adain}. 
We base the architecture for $D_\psi$ on that used in DC-GAN~\cite{DCGAN} except that we remove batch normalisation. 
Any deviations from this setup are described in the corresponding sections.

\paragraph{Training details}
We utilise the Adam optimiser~\cite{adam} with an initial learning rate of \num{2e-4} for $F_\theta$ and $G_\phi$, and \num{1e-5} for $D_\psi$. The learning rates are linearly decayed over time and we perform scheduled updates to train the networks. 
Additionally, the discriminator loss is regularised by using the $R_1$ zero-centered gradient penalty term~\cite{r1}. For a detailed list of hyperparameter values, see~\cref{sup:training}.

\begin{table*}[b!]
    \centering
    \begin{threeparttable}
    \addtolength{\tabcolsep}{-1pt}
      \begin{tabularx}{\textwidth}{Xcccc}
        \toprule
        Method                 &  GED $\downarrow$ (16) & GED $\downarrow$ (50) & GED $\downarrow$ (100) & HM-IoU $\uparrow$ (16) \\ 
        \midrule
        Kohl \etal (2018)~\cite{rel:probUnet}   & $0.320 \pm 0.030$ & \textemdash & $0.252 \pm \text{N/A}$\tnote{1} & $0.500 \pm 0.030$\\
        Kohl \etal (2019)~\cite{kohl2019hierarchical} & $0.270 \pm 0.010$ & \textemdash & \textemdash & $0.530 \pm 0.010$\\
        \midrule
        Hu \etal (2019)~\cite{multipleannotations}   & \textemdash & $0.267 \pm 0.012$ & \textemdash & \textemdash \\
        Baumgartner \etal (2019)~\cite{baumgartner2019phiseg} & \textemdash & \textemdash & $\mathbf{0.224 \pm} \text{\textbf{N/A}}$\tnote{2} & \textemdash \\
        Monteiro \etal (2020)~\cite{monteiro2020stochastic} & \textemdash & \textemdash & $0.225 \pm 0.002$\tnote{2} & \textemdash \\
        \midrule
        cGAN+$\mathcal{L}_\mathrm{ce}$ & $0.639 \pm 0.002$  & \textemdash & \textemdash & $0.477 \pm 0.004$ \\
        CAR (ours)
        & $\mathbf{0.264 \pm 0.002}$ & $\mathbf{0.248 \pm 0.004}$ & $0.243\pm 0.004$\tnote{2} & $\mathbf{0.592 \pm 0.005}$\\
        \bottomrule
      \end{tabularx}
      \begin{tablenotes}\footnotesize
      \item[1] This score is taken from~\cite{baumgartner2019phiseg}.
    \item[2] Note that when following the data split methodology used in~\cite{baumgartner2019phiseg,monteiro2020stochastic}
    and computing the GED (100) metric, we achieve a score of $0.228 \pm 0.009$ instead of $0.243 \pm 0.004$ (see~\cref{note:lidc_scores}).
      \end{tablenotes}
      \end{threeparttable}
      \caption{Mean GED and HM-IoU scores on LIDC. Top section: approaches using the original data splits defined by~\cite{rel:probUnet}, which we also adhere to; middle: approaches using random data splits; bottom: the $\mathcal{L}_\text{ce}$-regularised baseline and our CAR model.
      The three central columns show the GED score computed with 16, 50 and 100 samples, respectively. The rightmost column shows the HM-IoU score, computed with 16 samples. 
    The arrows $\uparrow$ and $\downarrow$~indicate~if~higher~or~lower~score~is~better.
    }
    \label{tab:lidc_comparison}
\end{table*}

\begin{figure*}[b!]%
    \centering
    \includegraphics[width=\textwidth]{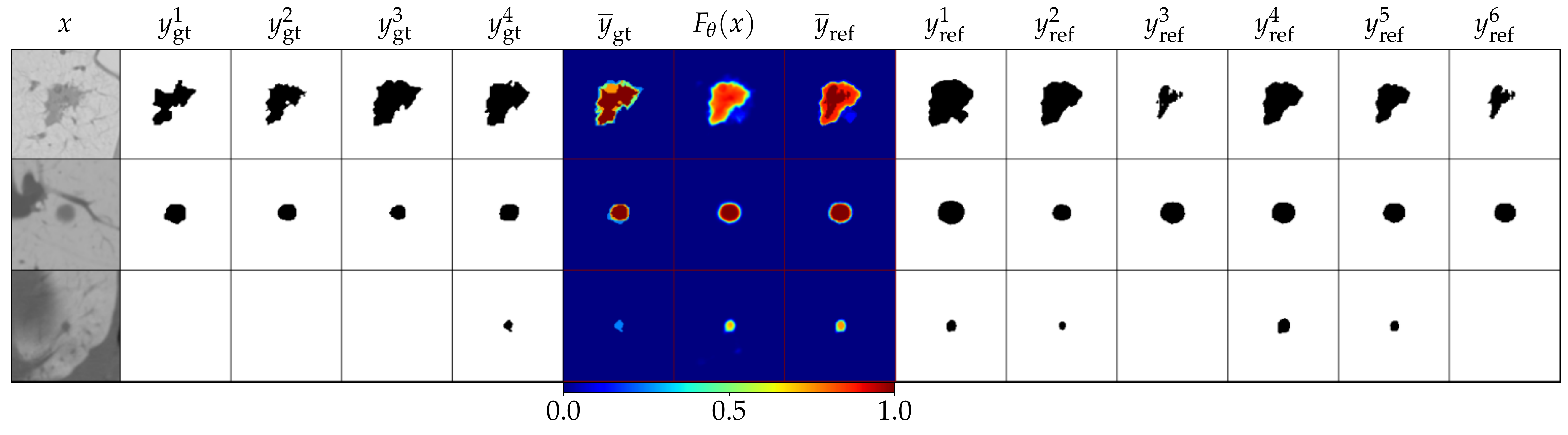}
    \caption{LIDC validation samples. From left to right: an input image $x$, followed by the four ground truth annotations $y_\text{gt}^1 \ldots y_\text{gt}^4$, the mean of the labels $\overline{y}_\text{gt}$, the output of the calibration network $F_\theta(x)$, the mean of the six refinement network samples $\overline{y}_\text{ref}$, shown in columns $y_\text{ref}^1 \ldots y_\text{ref}^6$.
    }%
    \label{fig:lidc_samples}%
\end{figure*}

\paragraph{Metrics} 
Following~\cite{rel:probUnet,kohl2019hierarchical,unsupervised_multimodal,baumgartner2019phiseg}, we use the Generalised Energy Distance (GED)~\cite{szekely2013energy} metric:
\begin{equation}
    \begin{aligned}
    D^2_\text{GED}(\p[\data], \q[\phi]) &= 2 \expc[s \sim \q[\phi], y \sim \p[\data]]{d(s, y)} 
    - \expc[s, s' \sim \q[\phi]]{d(s, s')} \\
    &\quad- \expc[y, y' \sim \p[\data]]{d(y, y')},
    \end{aligned}
    \label{eq:ged_metric_lidc_main}
\end{equation} 
where $d(s, y) = 1 - \mathrm{IoU}(s, y)$. 
As an additional metric, we follow~\cite{kohl2019hierarchical} in using the Hungarian-matched IoU (HM-IoU)~\cite{kuhn2005hungarian}. In contrast to GED, which naively computes diversity as $1-\mathrm{IoU}$ between all possible pairs of ground truth or sampled predictions, HM-IoU finds the optimal 1:1 matching between all labels and predictions, and therefore is more representative of how well the learnt predictive distribution fits the ground truth.

All experiments are performed in triplicate and we report results as mean and standard deviation. Further details regarding the exact implementation of the GED and HM-IoU metrics for each experiment can be found in~\cref{sup:lidc_details}.

\subsubsection{Learning shape diversity on the LIDC dataset}
\label{sec:shape_diversity}

To assess the accuracy and diversity of samples generated by our model, we use the Lung Image Database Consortium (LIDC)~\cite{data:lidc} dataset which consists of 1018 thoracic CT scans from 1010 lung cancer patients, graded independently by four expert annotators. We use the $180 \PLH 180$ crops from the preprocessed version of the LIDC dataset used and described in~\cite{rel:probUnet}. The dataset is split in 8882, 1996 and 1992 images in the training, validation and test sets respectively. All models are trained on lesion-centered $128 \PLH 128$ crops where at least one of the four annotations indicates a lesion. The final evaluation is performed on the provided test set.

In this task, we pretrain $F_\theta(x)$ with $\mathcal{L}_\mathrm{ce}$ in isolation, fix its weights, and then train $G_\phi$ with $\mathcal{L}_{\mathrm{G}}$, estimating $\mathcal{L}_\mathrm{cal}$ with 20 samples from $G_\phi$.
Note that we disclose further experiments with varying sample size in~\cref{tab:lidc_comaprison_n_samples} in~\cref{sup:more_lidc}. As a control, we train using the same architecture but replace $\mathcal{L}_\mathrm{cal}$ in the refinement network loss function with a cross entropy loss $\mathcal{L}_\mathrm{ce}$, as done in~\cite{luc2016semantic,elgan}. We denote this baseline throughout the manuscript as cGAN+$\mathcal{L}_\mathrm{ce}$.

\begin{figure*}[b!]
    \centering
    \subfloat[]{\includegraphics[height=4.7cm]{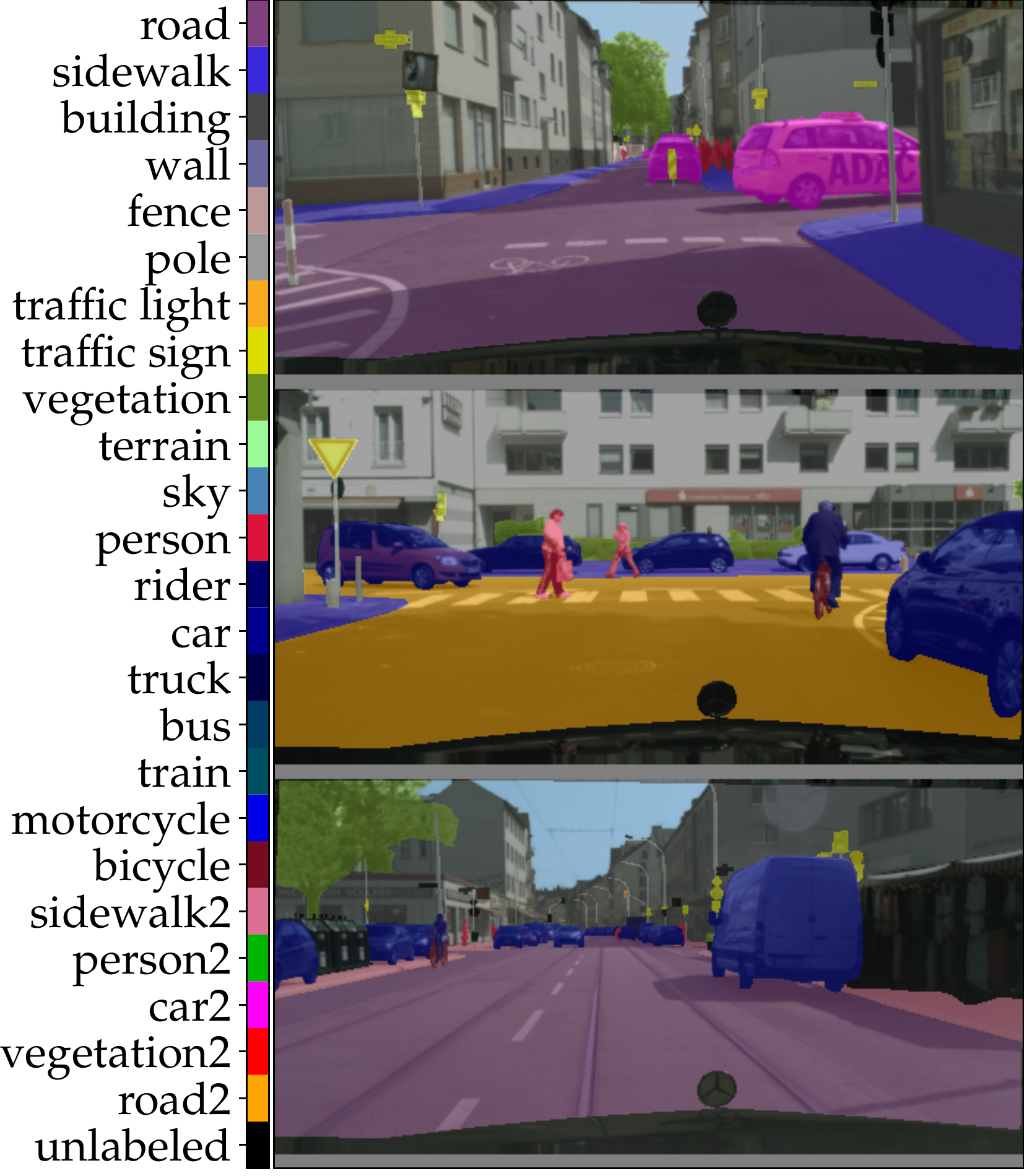}}
    \,
    \subfloat[]{\includegraphics[height=4.7cm]{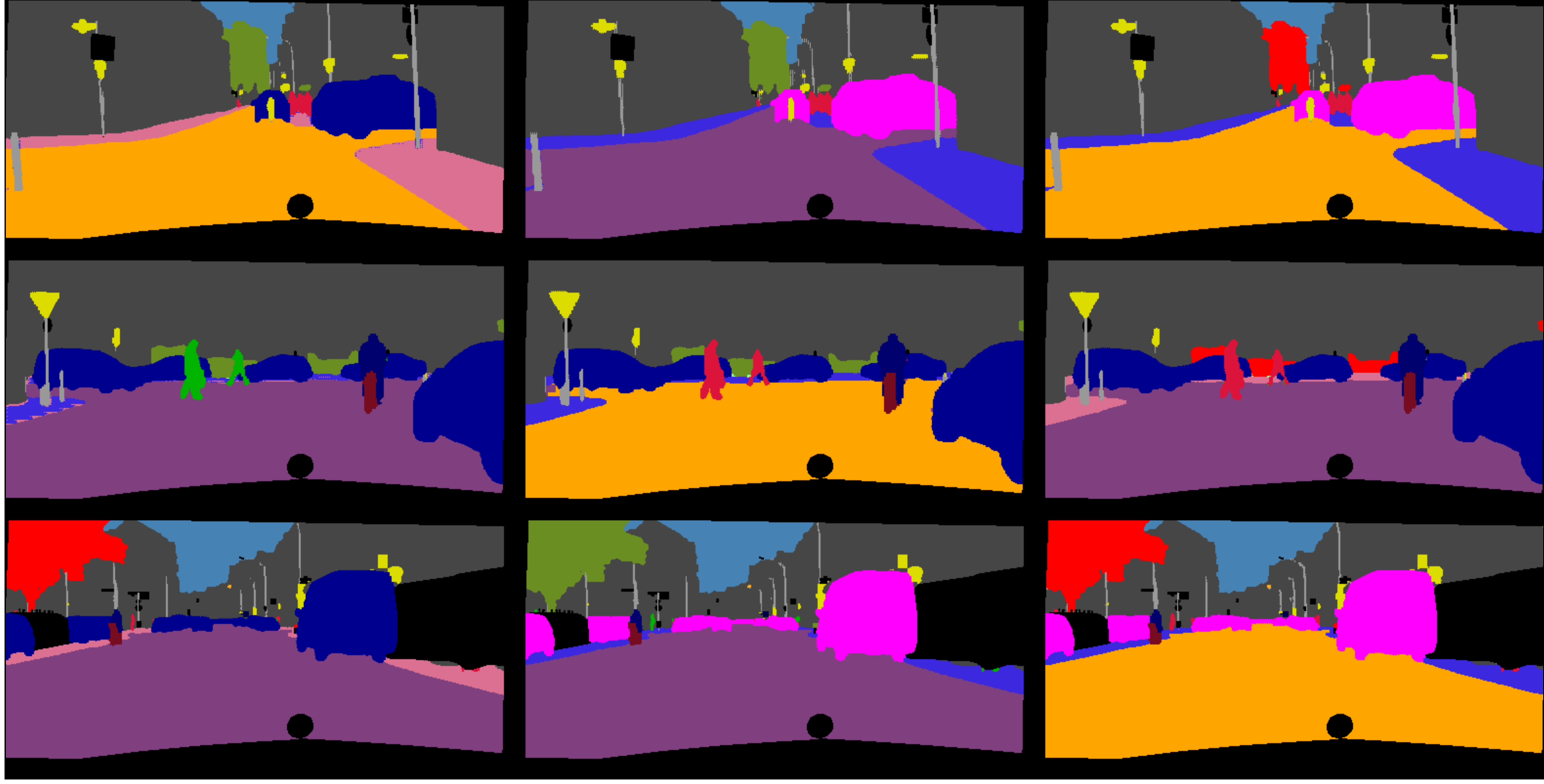}}
    \,
    \subfloat[]{\includegraphics[height=4.7cm]{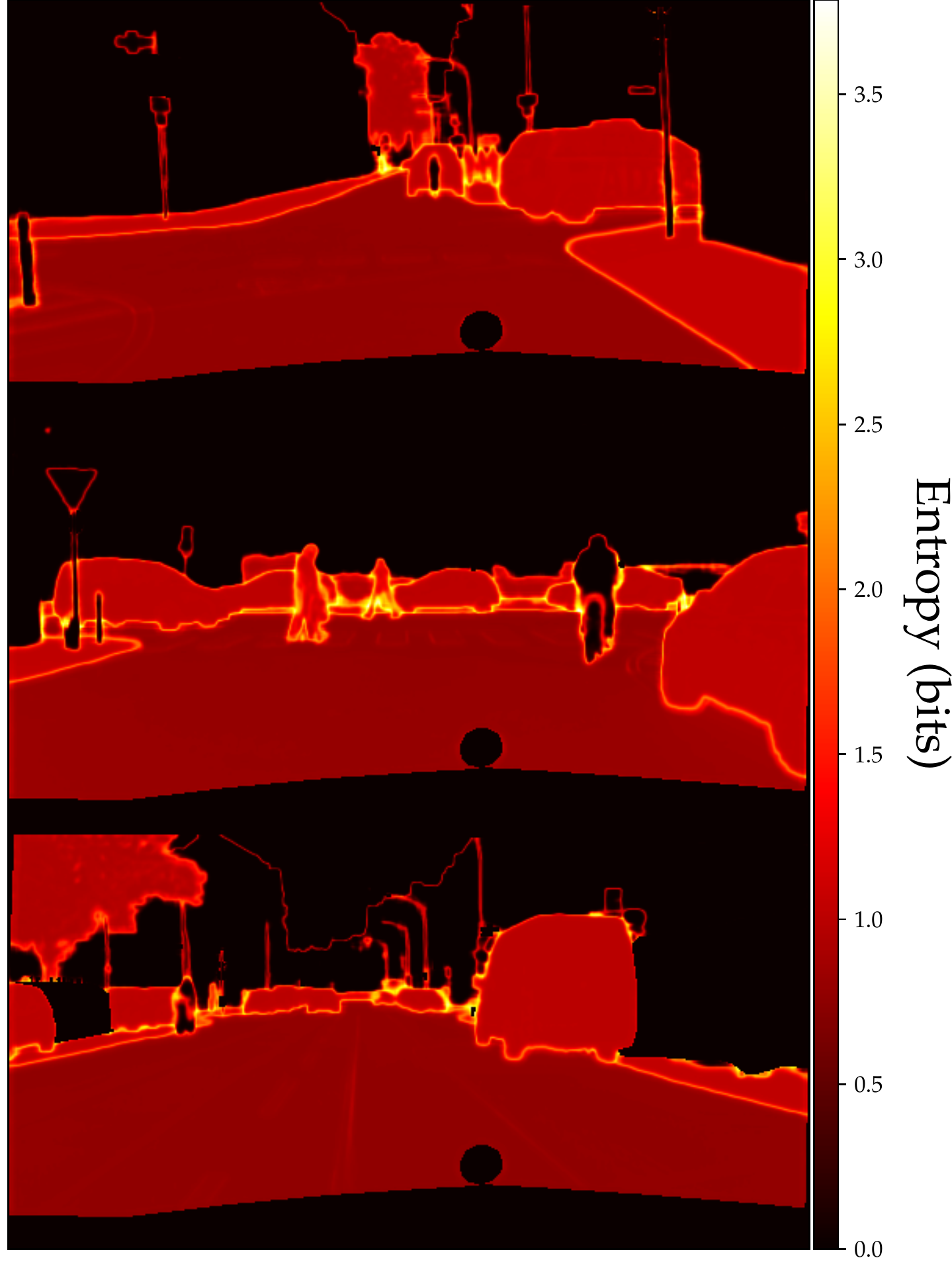}}
    \caption{\textbf{(a)} Input images overlaid with the corresponding labels. \textbf{(b)} Samples obtained from the refinement network. \textbf{(c)} Aleatoric uncertainty computed as the entropy of the calibration output.}%
    \label{fig:cityscapes_samples}%
\end{figure*}

Our results show that the CAR model performs on par with other state-of-the-art methods \wrt the GED score, and outperforms them on the HM-IoU score (only available for~\cite{rel:probUnet,kohl2019hierarchical}). Numerical results are summarised in~\cref{tab:lidc_comparison} and the diversity and fidelity of sampled predictions are illustrated in~\cref{fig:lidc_samples}. In contrast, the $\mathcal{L}_\mathrm{ce}$-regularised baseline collapses the predictive distribution, showing no perceptible diversity between samples (see~\cref{fig:lidc_samples_vae}b in~\cref{{sup:lidc_lvm_exps}}), which results in a stark increase in the mean GED score, and decrease in the HM-IoU score. 

\subsubsection{Learning a calibrated distribution on a multimodal Cityscapes dataset}
\label{sec:evaluation_cs}

The Cityscapes dataset contains $1024 \PLH 2048$ RGB images of urban scenes, and corresponding segmentation maps. It consists of 2975 training, 500 validation and 1525 test images. Following~\cite{rel:probUnet}, we use a stochastic version of the Cityscapes dataset with 19 semantic classes, downsample images and segmentation maps to a spatial resolution of $256 \PLH 512$, and report results on the validation set. Controlled multimodality is established by augmenting the dataset with 5 new classes: \textit{sidewalk2}, \textit{person2}, \textit{car2}, \textit{vegetation2} and \textit{road2}, introduced by flipping their original counterparts with probabilities $\nicefrac{8}{17}$, $\nicefrac{7}{17}$, $\nicefrac{6}{17}$, $\nicefrac{5}{17}$ and $\nicefrac{4}{17}$, respectively (see~\cref{fig:cityscapes_samples}a), giving the dataset a total of 24 semantic classes. 

To demonstrate that our approach can be easily integrated on top of any existing black-box segmentation model $B$, we employ the network from~\cite{black_box_cs}, trained on the official Cityscapes dataset, achieving a mIoU of 0.79 on the test set. We utilise its predictions as input to a smaller version of our calibration network $F_\theta$ comprising 5 convolutional blocks, each composed of a $3 \PLH 3$ convolutional layer followed by batch normalisation, a leaky ReLU activation, and a dropout layer with 0.1 dropout rate.
We pretrain the calibration network in isolation, and subsequently apply it in inference mode while adversarially training the refinement network. We use a batch size of 16, and train with $\mathcal{L}_{\mathrm{G}}$, estimating $\mathcal{L}_\mathrm{cal}$ with 7 samples from $G_\phi$.
The same baseline from the LIDC experiment is employed, where $\mathcal{L}_\mathrm{cal}$ is replaced with $\mathcal{L}_\mathrm{ce}$. 
As a second control experiment, we completely omit the calibration network and instead condition the refinement network on the known ground truth pixelwise categorical distribution over the label, representing the ideal output of $F_{\theta}$. This baseline allows us to directly evaluate the quality of sampling administered from the refinement network, and we denote it as cGAN+$\mathcal{L}_\mathrm{cal}$~(ground truth).

When training the refinement network with a cross entropy loss instead of the calibration loss $\mathcal{L}_\mathrm{cal}$, the predictive distribution collapses, making the output deterministic. Conversely, when we train our refinement network with $\mathcal{L}_\mathrm{cal}$, the learnt predictive distribution is well adjusted, with high diversity and reconstruction quality, significantly outperforming the current state-of-the-art, as shown in~\cref{tab:cs_comparison}. \cref{fig:cityscapes_samples}b displays representative sampled predictions from our model for three input images and~\cref{fig:cityscapes_samples}c illustrates the corresponding aleatoric uncertainty maps extracted from $F_\theta(x)$. The learnt multimodality and noise in the dataset are reflected by regions of high entropy, where objects belonging to the different stochastic classes consistently display distinct shades of red, corresponding to their respective flip probabilities. Finally, we show that when using the ground truth distribution as the input and calibration target to the refinement network, we attain an almost perfect GED score ($0.038~\pm~0.00$). 

\begin{figure}[t!]
        \centering
        \includegraphics[width=0.45\textwidth]{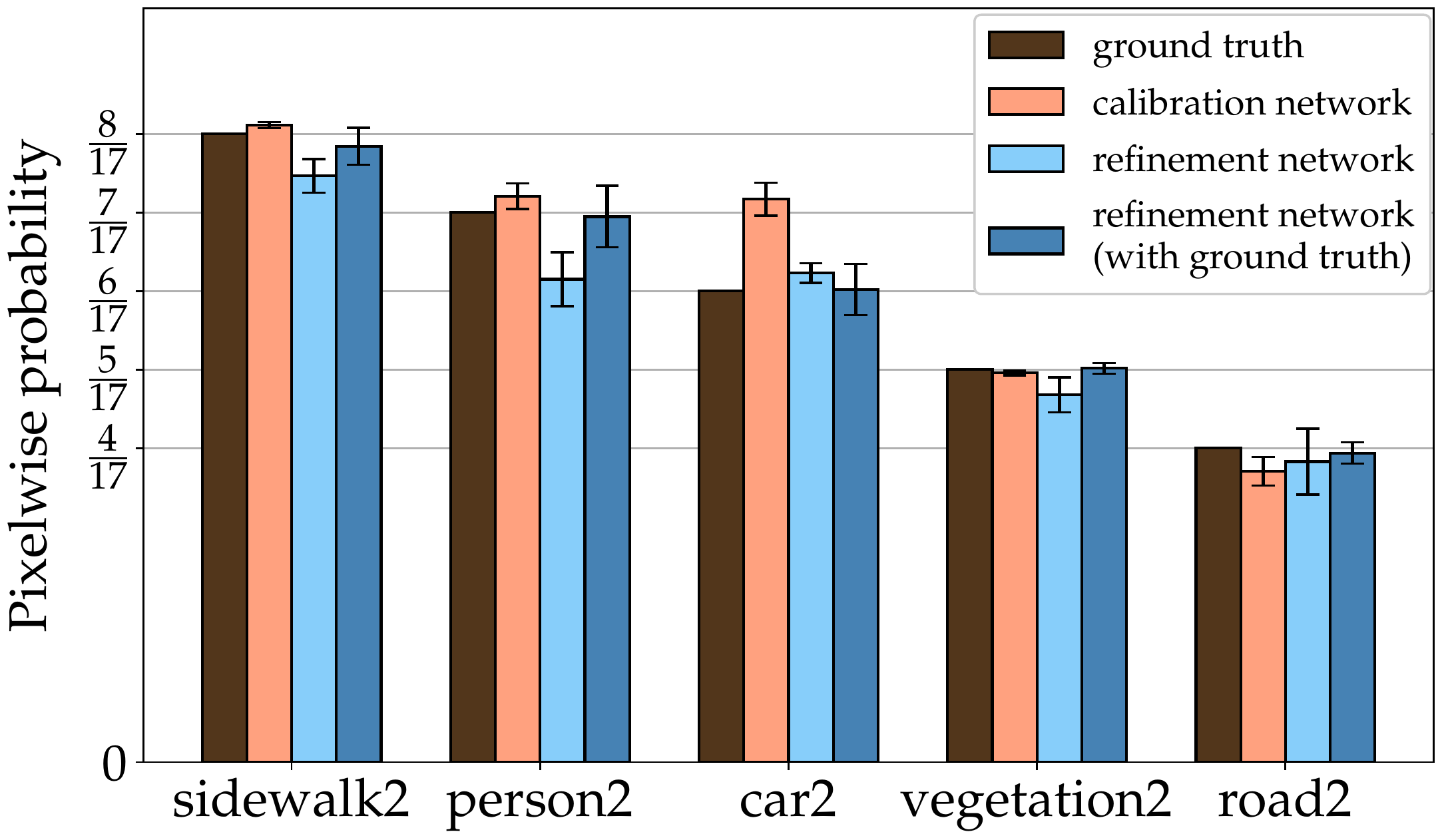}
        \caption{Calibration of the pixelwise probabilities of the five stochastic classes. Note that the calibration network (in orange) is conditioned on black-box predictions.}
        \label{fig:cs_calibration}
\end{figure}

Since we manually set the flip probabilities for each stochastic class in this dataset, we can directly assess the calibration of our model by comparing the ground truth probabilities to the predicted probabilities from the calibration or refinement network. For $F_\theta$ we use the mean confidence values for each class, and for $G_\phi$ we obtain the empirical mean class probabilities via $\overline{G}_{\phi}(F_\theta(x))$, computed from 16 samples (see~\cref{sup:cs_details} for more details).
The ensuing results are shown graphically in~\cref{fig:cs_calibration}, which illustrates the calibration of our models on the stochastic classes,
evaluated over the entire dataset. This demonstrates that our models are well calibrated, with the calibration offset, computed as the absolute difference between the ground truth and predicted probabilities, being approximately $6\%$ in the worst case (class "car2" for the calibration network). Note that the average calibration offset for $F_\theta(x)$ across the stochastic classes is $1.6\%$. Further, the ground truth conditioned baseline is almost perfectly calibrated, in accord to the near-optimal GED score reported in~\cref{tab:cs_comparison}. Thus, we demonstrate that $G_\phi$ successfully learns calibrated refinement of the predictions from $F_\theta$, where the quality of the final predictive distribution depends on the quality of $F_\theta(x)$.

\begin{table}
    \centering
    \begin{tabular}{lc}
        \toprule
        Method  & GED     \\ 
        \midrule
        Kohl \etal (2018)~\cite{rel:probUnet} & $0.206 \pm \text{N/A}$  \\
        \midrule
        cGAN+$\mathcal{L}_\mathrm{ce}$ & $0.632 \pm 0.07$  \\
        CAR (ours)& $\mathbf{0.164 \pm 0.01}$ \\
        \midrule
        cGAN+$\mathcal{L}_\mathrm{cal}$ (ground truth)   & $0.038 \pm 0.00$   \\
        \bottomrule
    \end{tabular}
    \caption{Mean GED scores on the modified Cityscapes. Top section: competing model; middle: 
    $\mathcal{L}_\mathrm{ce}$-regularised baseline and CAR model; bottom: ground truth calibrated refinement network (cGAN). GED scores are computed using 16 samples.}
    \label{tab:cs_comparison}
\end{table}

\begin{figure}[h]
  \centering
    \includegraphics[width=0.395\textwidth]{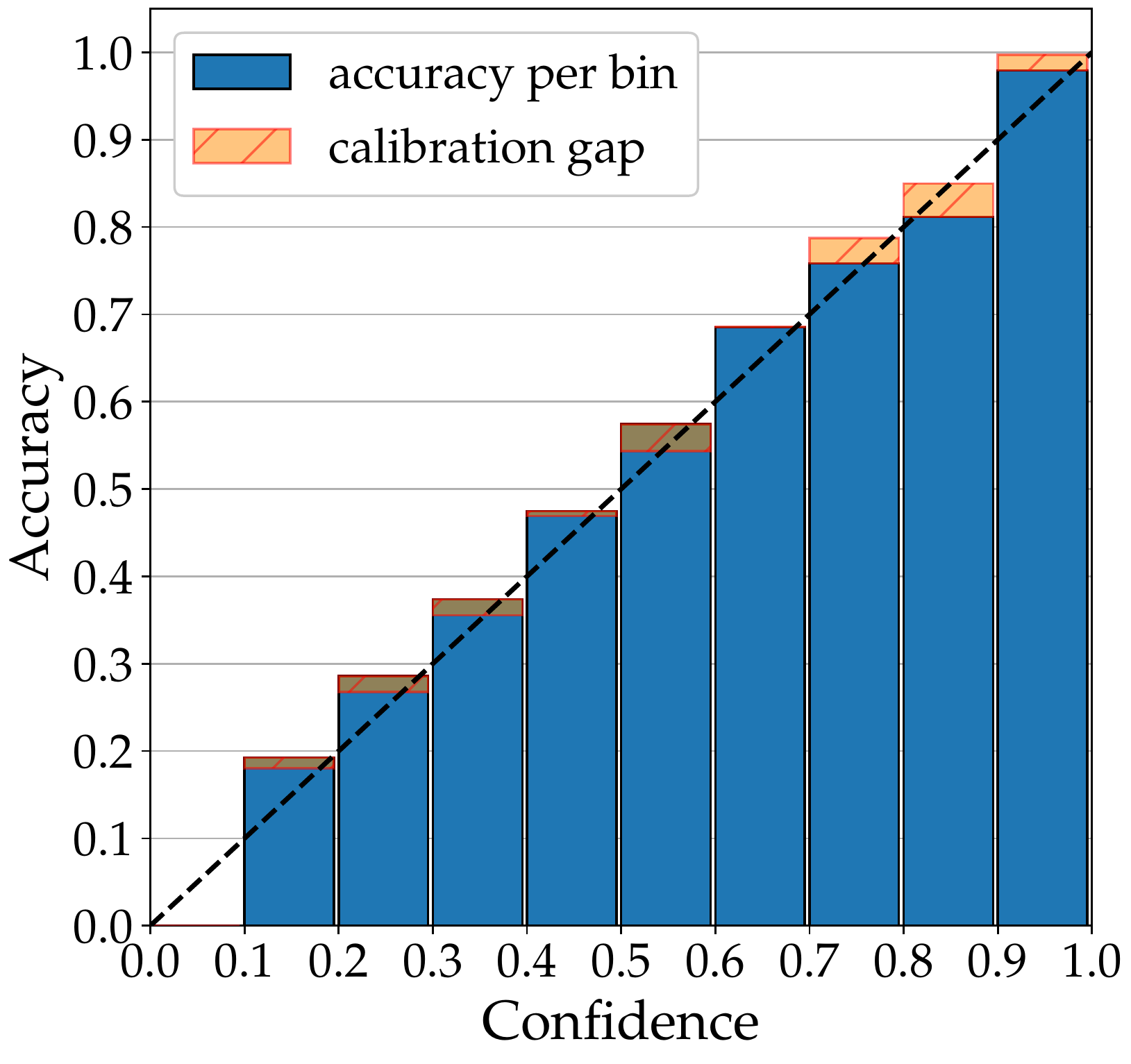}
  \caption{Reliability diagram for the calibration network. The corresponding ECE score is $2.15\%$.}%
  \label{fig:cs_reliability}
\end{figure}

In order to further scrutinise the calibration quality of $F_\theta(x)$, we construct a reliability diagram and compute the corresponding expected calibration error (ECE), following~\cite{calibration}. To create the diagram, each pixel is considered independently, and the class confidences are binned into 10 equal intervals of size 0.1. We then compute the accuracy for all predictions in each bin. \cref{fig:cs_reliability} shows the reliability diagram for the calibration network, where the orange bars depict the calibration gap, defined as the difference between the mean confidence for each interval and the corresponding accuracy. 
The corresponding ECE score amounts to $2.15\%$. Note that this also considers the average calibration error computed for the stochastic classes, where we randomly sample the labels according to the predefined probabilities. Hence, we confirm that $F_\theta(x)$ is well calibrated.

An important outstanding issue in our approach is that the calibration network may not perfectly capture the class-probabilities, \eg as seen from the \textit{car2} category in~\cref{fig:cs_calibration}. Since we calibrate the refinement network relative to the calibration target provided by $F_\theta$, such errors can be propagated into $G_\phi$. Calibration issues in modern neural networks are well documented~\cite{calibration,kull2019beyond,zhang2020mix}---these have been attributed to several factors, such as long-tailed data distributions, out-of-distribution inputs, specific network architectural elements or optimising procedures. Even though we did not find it necessary, ad hoc solutions to miscalibration, such as temperature scaling~\cite{hinton2015distilling}, Dirichlet calibration~\cite{kull2019beyond} etc., can be readily applied to the calibration network to improve the predicted probabilities, and thus the overall calibration of our model. Therefore, an important direction for future work is towards improving the calibration of deep neural networks.

\section{Conclusion}
In this work, we developed a novel framework for semantic segmentation capable of learning a calibrated multimodal predictive distribution, closely matching the ground truth distribution of labels. We attained improved results on a modified Cityscapes dataset and competitive scores on the LIDC dataset, indicating the utility of our approach on real-world datasets. 
We also showed that our approach can be easily integrated
into any off-the-shelf, deterministic, black-box semantic segmentation model, enabling sampling an arbitrary number of plausible segmentation maps. By highlighting regions of high data uncertainty and providing multiple valid label proposals, our approach can be used to identify and resolve ambiguities, diminishing risk in safety-critical systems. Therefore, we expect our approach to be particularly beneficial for applications such as map making for autonomous driving or computer-assisted medical diagnostics.
Finally, even though the primary focus of this work is semantic segmentation, we demonstrated the versatility of our method with an illustrative toy regression problem, alluding to a broader applicability beyond semantic segmentation.

\section{Acknowledgements}
The authors would like to thank Mohsen Ghafoorian, Animesh Karnewar, Michael Hofmann, Marco Manfredi and Atanas Mirchev for their invaluable feedback and support.

\small

{\small
\bibliographystyle{ieee_fullname}
\bibliography{refs}
}

\newpage
\onecolumn 

\clearpage
\pagenumbering{arabic}
\setcounter{page}{1}

\begin{appendices}

\section{Implementation details}

In this section we describe the overall training procedure and delve into the training and evaluation details for the stochastic segmentation experiments on the LIDC dataset and the modified Cityscapes dataset.

\subsection{Training procedure}
\label[supplementary]{sup:training}

\cref{alg:training} outlines the practical procedure used to pretrain the calibration network, and the subsequent training of the refinement network. Even though the two networks can be trained end-to-end at once, in our experiments we use the two-step training procedure to stabilise the training and reduce the memory consumption on the GPU. This way we are able to fit larger batches and/or more samples for the estimate of $\mathcal{L}_\text{cal}$. \cref{alg:inference} shows the inference procedure for obtaining $M$ output samples.

\begin{algorithm}[!h]
	\caption{Model training with calibration network pretraining}
	\hspace*{\algorithmicindent} \textbf{require:} training data $\data$, number of samples $M$, learning rate $\eta$, calibration loss scale $\lambda$;
	\begin{algorithmic}[1]
	    \Procedure{Training}{$\data,M,\eta,\lambda$}
    	    \While {not converged}  \Comment{Pretraining of $F_\theta$}
    	        \State Sample batch $\{x, y\}_t \in \data$
    	        \State Update $\theta_{t+1}$ with $-\eta\nabla_{\theta_t}\mathcal{L}_{\mathrm{ce}}(\{x, y\}_t, \theta_t)$ 
    	    \EndWhile
    	    \While {not converged}  \Comment{Adversarial training of $G_\phi$ and $D_\psi$}
    	        \State Sample batch $\{x, y\}_t \in \data$
    	        \For {$i=1,2,\ldots,M$}
        			\State Sample $y^{i,t}_\text{ref} = G_{\phi_t}(F_{\theta^*}(x_t), \epsilon_i) \quad \mathrm{where} \; \epsilon_i \sim \gauss{0, 1}$
        		\EndFor
    	        \State Compute $\overline{G}_{\phi}(F_\theta(x))^{t} = \frac{1}{M}\sum_{i=1}^M y^{i,t}_\text{ref}$
    	        \State Compute $\mathcal{L}_{\mathrm{cal}}(x_t, \theta^*, \phi_t) = \sum_{i,j,k}
    	        \left(\overline{G}_{\phi}(F_\theta(x))^{t}\left(\log{\overline{G}_{\phi}(F_\theta(x))^{t}} - \log{F_{\theta^*}(x_t)}\right)\right)_{i,j,k}$
    	        \State Update $\phi_{t+1}$ with $-\eta\nabla_{\phi_t}\left(\mathcal{L}_{\mathrm{G}}(\theta^*, \phi_t, \{x, y\}_t) + \lambda\mathcal{L}_{\mathrm{cal}}(x_t, \theta^*, \phi_t)\right)$
    	        \State Update $\psi_{t+1}$ with $-\eta\nabla_{\psi_t}\mathcal{L}_{\mathrm{D}}(\theta^*, \phi_t, \psi_t, \{x, y\}_t)$
    	    \EndWhile
	    \EndProcedure
	\end{algorithmic} 
    \label{alg:training}
\end{algorithm}

\begin{algorithm}
	\caption{Inference procedure} 
	\hspace*{\algorithmicindent} \textbf{require:} test data point $x$, number of samples $M$;
	\begin{algorithmic}[1]
	    \Procedure{Inference}{$x,M$} \Comment{Using $\theta^*$ and $\phi^*$ from \cref{alg:training}}
            \For {$i=1,2,\ldots,M$} 
    			\State Sample $y^i_\text{ref} = G_{\phi^*}(F_{\theta^*}(x), \epsilon_i) \quad \mathrm{where} \; \epsilon_i \sim \gauss{0, 1}$
    		\EndFor
		\EndProcedure
	\end{algorithmic} 
    \label{alg:inference}
\end{algorithm}

Notice that any off-the-shelf optimisation algorithm can be used to update the parameters $\theta$, $\phi$ and $\psi$. For the segmentation experiments, we utilise the Adam optimiser~\cite{adam} with $\beta_1 = 0.5$, $\beta_2 = 0.99$ and weight decay of \num{5e-4}. $F_\theta$ is trained with a learning rate of \num{2e-4} which is then lowered to \num{1e-4} after 30 epochs. $G_\phi$ and $D_\psi$ are updated according to a schedule, where $G_\phi$ is updated at every iteration, and $D_\psi$ is trained in cycles of 50 iterations of weight updating, followed by 200 iterations with fixed weights. The refinement network is trained with an initial learning rate of \num{2e-4}, lowered to \num{1e-4} after 30 epochs, whereas the discriminator has an initial learning rate of \num{1e-5}, lowered to \num{5e-6} after 30 epochs. Additionally, we utilise the $R_1$ zero-centered gradient penalty term~\cite{r1}, to regularise the discriminator gradient on real data with a weight of 10. Other hyperparameter specifics such as the batch-size and whether we inject stochasticity via random noise samples or latent code samples, depend on the experiment and are disclosed in the respective sections below or in the main text.

\subsection{1D bimodal regression}
\label[supplementary]{sup:regression_details}
In the following we derive the mean squared error form of the cross entropy and calibration losses used in experiment~\cref{sec:toy_regression} under the assumption that the likelihood model for $\q[\theta]$ and $\q[\phi]$ is a univariate Gaussian distribution with a fixed unit variance. Using the setup from~\cref{eq:loss_ce} it then follows that:
\begin{align}
    \mathcal{L}_{\mathrm{ce}}(\data, \theta) &= -\expc[\p[\data]{x,y}]{\log \gauss{y}{F_\theta(x),\,1}} \\
    &= \frac{1}{2}\expc[\p[\data]{x,y}]{\left(y - F_\theta(x)\right)^2} + \mathrm{const}.
\label{eq:derivation_toy_lce}
\end{align}
Based on the definition of the calibration loss in~\cref{eq:loss_cal} we show that:
\begin{align}
    \mathcal{L}_{\mathrm{cal}}(\data, \theta, \phi) &= \expc[\p[\data]]{\kl{\gauss{y}{\overline{G}_{\phi}(F_\theta(x)),\,1}}{\gauss{y}{F_\theta(x),\,1}}} \\
    &= \expc[\p[\data]]{\expc[\gauss{y}{\overline{G}_{\phi}(F_\theta(x)),\,1}]{\log \gauss{y}{\overline{G}_{\phi}(F_\theta(x)),\,1} - \log \gauss{y}{F_\theta(x),\,1}}} \\
    &= \expc[\p[\data]]{\expc[\gauss{y}{\overline{G}_{\phi}(F_\theta(x)),\,1}]{-\frac{1}{2}\left(y - \overline{G}_{\phi}(F_\theta(x))\right)^2 + \frac{1}{2}\left(y- F_\theta(x)\right)^2}} + \mathrm{const} \\
    &= \expc[\p[\data]]{\expc[\gauss{y}{\overline{G}_{\phi}(F_\theta(x)),\,1}]{y\overline{G}_{\phi}(F_\theta(x)) -\frac{1}{2}\overline{G}_{\phi}(F_\theta(x))^2 - yF_\theta(x) + \frac{1}{2}F_\theta(x)^2}} + \mathrm{const} \\
    &= \expc[\p[\data]]{\overline{G}_{\phi}(F_\theta(x))^2 -\frac{1}{2}\overline{G}_{\phi}(F_\theta(x))^2 - \overline{G}_{\phi}(F_\theta(x))F_\theta(x) + \frac{1}{2}F_\theta(x)^2} + \mathrm{const} \\
    &= \frac{1}{2}\expc[\p[\data]]{\left(\overline{G}_{\phi}(F_\theta(x)) - F_\theta(x)\right)^2} + \mathrm{const}.
\label{eq:derivation_toy_lcal}
\end{align}

\subsection{LIDC}
\label[supplementary]{sup:lidc_details}

\paragraph{Architectures} For the calibration network, $F_\theta$, we use the encoder-decoder architecture from SegNet~\cite{segnet}, with a softmax activation on the output layer.

\paragraph{Training}  

During training, we draw random  $180 \PLH 180$ image-annotation pairs, and we apply random horizontal flips and crop the  data to produce $128 \PLH 128$ lesion-centered image tiles. All of our models were implemented in PyTorch and trained for 80k iterations on a single 32GB Tesla V100 GPU.

We train all our models for the LIDC experiments using 8-dimensional noise vectors in the cGAN experiments, or latent codes in the cVAE-GAN experiments. This value was empirically found to perform well, sufficiently capturing the shape diversity in the dataset. Additionally, in the refinement networks loss, we set the weighting parameter $\lambda$ in the total generator loss, defined in~\cref{eq:loss_total} in the main text, so as to establish a ratio of $\mathcal{L}_{\mathrm{G}}:\mathcal{L}_{\mathrm{cal}} = 1:0.5$, where $\mathcal{L}_{\mathrm{G}}$ is the adversarial component of the loss, and $\mathcal{L}_{\mathrm{cal}}$ is the calibration loss component. In practice, the actual weights used are 10 for $\mathcal{L}_{\mathrm{G}}$, and 5 for $\mathcal{L}_{\mathrm{cal}}$.

\paragraph{Evaluation} 

Following~\cite{rel:probUnet}, \cite{kohl2019hierarchical}, \cite{unsupervised_multimodal}, and~\cite{baumgartner2019phiseg} we use the Generalised Energy Distance (GED)~\cite{szekely2013energy} metric, given as:

\begin{equation}
    D^2_\text{GED}(\p[\data], \q[\phi]) = 2 \expc[s \sim \q[\phi], y \sim \p[\data]]{d(s, y)} - \expc[s, s' \sim \q[\phi]]{d(s, s')} - \expc[y, y' \sim \p[\data]]{d(y, y')},
    \label{eq:ged_metric_lidc}
\end{equation} 

where $d(s, y) = 1 - \mathrm{IoU}(s, y)$. Intuitively, the first term of~\cref{eq:ged_metric_lidc} quantifies the disparity between sampled predictions and the ground truth labels, the second term---the diversity between the predictions, and the third term---the diversity between the ground truth labels. It is important to note that the GED is a sample-based metric, and therefore the quality of the score scales with the number of samples. We approximate the expectations with all 4 ground truth labels ($y \sim \p[\data])$ and 16, 50 or 100 samples from the model ($s \sim \q_\phi$) for each input image $x$. 

As~\cite{kohl2019hierarchical} have pointed out, even though the GED metric is a good indicator for how well the learnt predictive distribution fits a multimodal ground truth distribution, it can reward high diversity in sampled predictions even if the individual samples do not show high fidelity to the ground truth distribution. As an alternative metric that is less sensitive to such degenerate cases,~\cite{kohl2019hierarchical} propose to use the Hungarian-matched IoU (HM-IoU), which finds the optimal 1:1 IoU matching between ground truth samples and sampled predictions. Following~\cite{kohl2019hierarchical}, we duplicate the set of ground truth labels so that the number of both ground truth and predicted samples are 16, and we report the HM-IoU as the average of the best matched pairs for each input image.

In the main text we show the evaluated performance with both GED and HM-IoU metrics over the entire test set and compute the IoU on only the foreground of the sampled labels and predictions. 
In the case where both the matched up label and prediction do not show a lesion, the IoU is set to 1, so that a correct prediction of the absence of a lesion is rewarded.

\label{note:lidc_scores}
Note that the methods of~\cite{baumgartner2019phiseg}, \cite{monteiro2020stochastic}
and~\cite{multipleannotations}, whose GED scores we report in~\cref{tab:lidc_comparison}, use test sets that differ from the original splits defined in~\cite{rel:probUnet}, which are used in~\cite{rel:probUnet,kohl2019hierarchical} and our work. \cite{baumgartner2019phiseg} and \cite{monteiro2020stochastic} 
uses a random 60:20:20 split for the training, testing and validation sets, and 100 samples to compute the GED score, whereas~\cite{multipleannotations} use a random 70:15:15 split and 50 samples to compute the GED score. Due to the lack of reproducibility, we do not consider this the conventional way of benchmarking. Therefore, in~\cref{tab:lidc_comparison} we only report the scores for our models evaluated on the original splits. Nevertheless, we also trained and tested our  
CAR model on the split methodology defined by~\cite{baumgartner2019phiseg}, and also used by~\cite{monteiro2020stochastic},
to enable a fairer comparison. This improved our GED score from $0.243\pm 0.004$ to $0.228 \pm 0.009$, while the HM-IoU, evaluated using 16 samples, remained similar at $0.590 \pm 0.007$ (in the original splits we achieved an HM-IoU score of $0.592 \pm 0.005$). This shows that different random splits can significantly affect the final performance \wrt GED score, while HM-IoU appears to be a more robust metric.

\subsection{Cityscapes}
\label[supplementary]{sup:cs_details}

\paragraph{Architectures} For the calibration network $F_\theta$, we design a small neural network with 5 convolutional blocks, each comprised of a $3 \PLH 3$ convolutional layer, followed by a batchnorm layer and a leaky ReLU activation. The network is activated with a softmax function.

\paragraph{Training} 

During training, we apply random horizontal flips, scaling and crops of size $128 \PLH 128$ on the image-label pairs. All of our models were implemented in PyTorch and trained for 120k training iterations on a single 16GB Tesla V100 GPU.

We train all our models for the modified Cityscapes experiments using 32-dimensional noise vectors. Similarly to the LIDC experiments, this value was empirically found to perform well, however, it can be further tuned. As commonly practiced, we use the ignore-masks provided by the Cityscapes dataset to filter out the cross entropy, calibration and adversarial losses during training on the unlabelled pixels. Similarly to our LIDC experiment, we use a weight of 10 for $\mathcal{L}_{\mathrm{G}}$, and 5 for $\mathcal{L}_{\mathrm{cal}}$ in the refinement networks loss.

\paragraph{Evaluation} The GED metric for Cityscapes is implemented as described in the appendix of~\cite{rel:probUnet} and evaluated across the entire validation set. In this dataset we have full knowledge of the ground truth class distribution and therefore we compute the GED metric by using the probabilities of each mode directly, as follows:

\begin{equation}
    D^2_\text{GED}(\p[\data], \q[\phi]) = 2 \expc[s \sim \q[\phi], y \sim \p[\data]]{d(s, y)w(y)} - \expc[s, s' \sim \q[\phi]]{d(s, s')} - \expc[y, y' \sim \p[\data]]{d(y, y')w(y)w(y')},
    \label{eq:ged_metric_cs}
\end{equation} 

where $w(\cdot)$ is a function mapping the mode of a given label $y$ to its corresponding probability mass. The distance $d(s,y)$ is computed using the average IoU of the 10 switchable classes only, as done in~\cite{rel:probUnet}. In the cases where none of the switchable classes are present in both the ground truth label and the prediction paired up in $d(s,y)$, the distance score is not considered in the expectation. We use 16 samples to compute the GED score.

For the calibration results presented in~\cref{fig:cs_calibration}, \cref{sec:evaluation_cs} in the main text, we compute the calibration network class-probabilities using the raw predictions of $F_\theta(x)$. We obtain class masks by computing the overlap between the ground truth labels and the black-box predictions for each class. Using these masks we then compute the average class-wise probabilities.
The probabilities for the refinement network $G_\phi$ were computed as the average over 16 samples. Here the class masks are obtained by finding the pixels that are specified as the class of interest in the ground truth labels.

\newpage
\section{Additional experiment results}

To reinforce the results reported in~\cref{sec:evaluation} we present supplementary results for the bimodal regression experiment and the LIDC and Cityscapes segmentation experiments. 

\subsection{1D bimodal regression}
\label[supplementary]{sup:more_regression}

\begin{figure}[!ht]%
    \centering
    \subfloat[]{\includegraphics[width=0.43\textwidth]{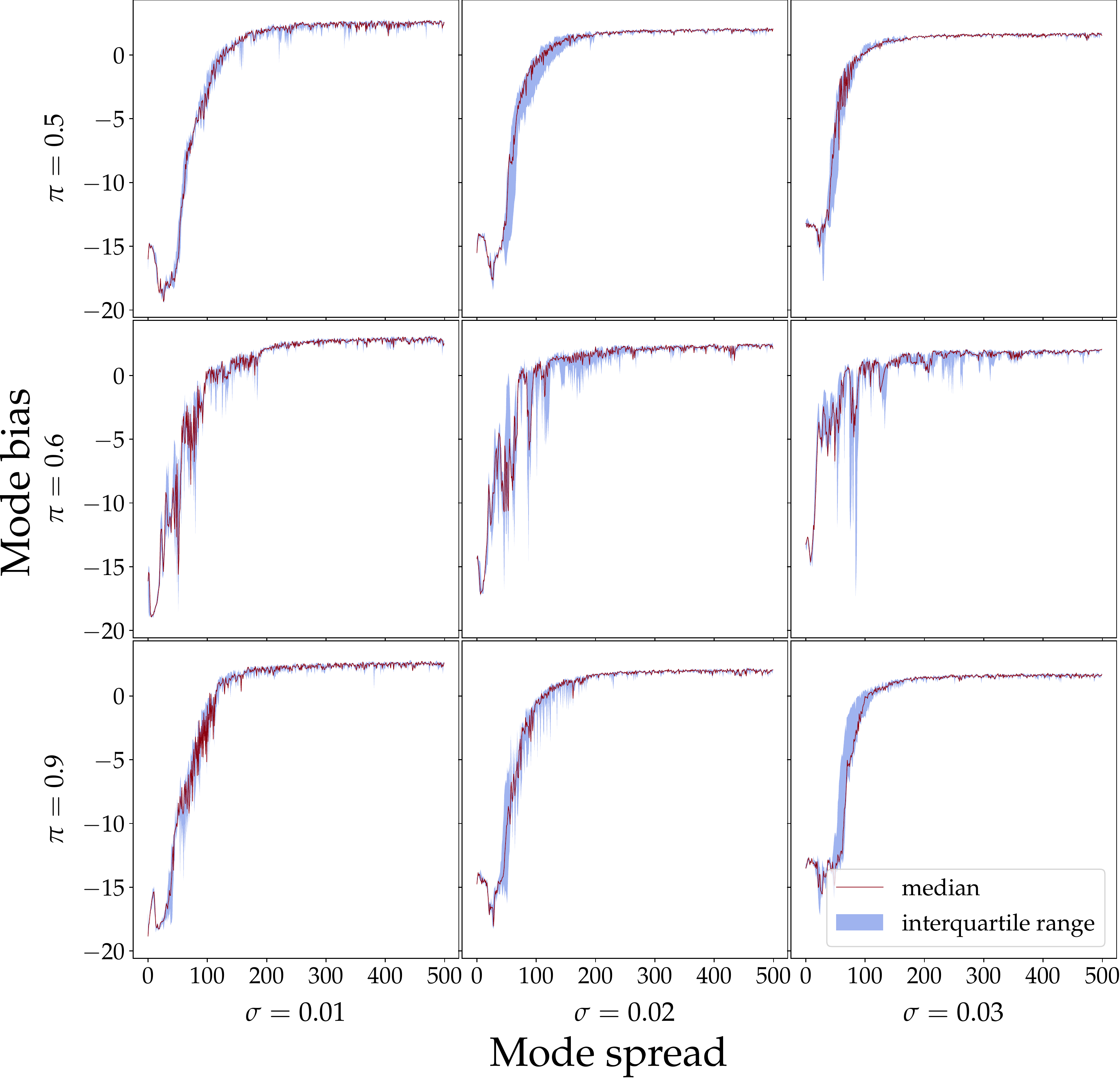}}
    \,
    \subfloat[]{\includegraphics[width=0.43\textwidth]{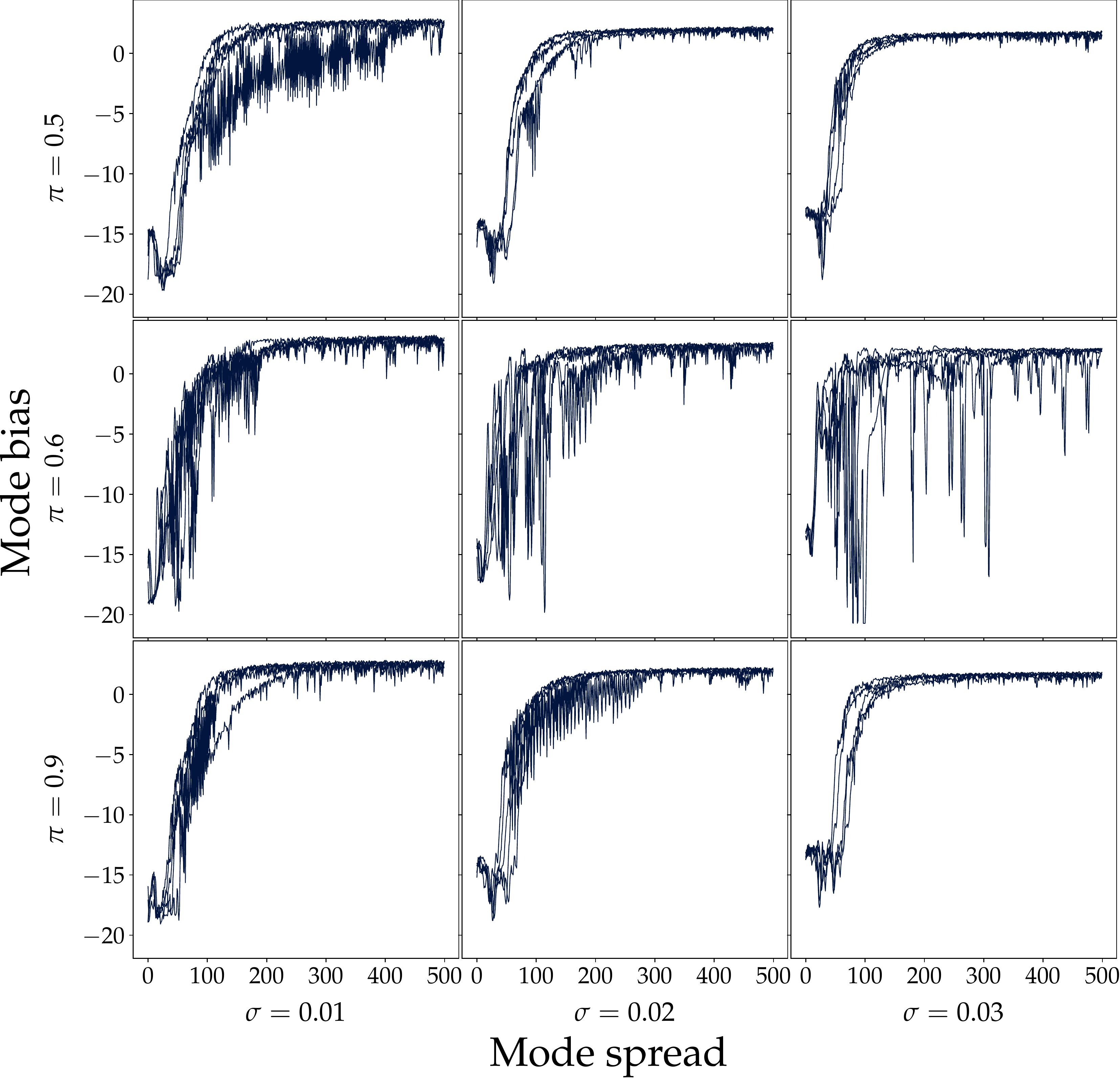}}
    \\
    \subfloat[]{\includegraphics[width=0.43\textwidth]{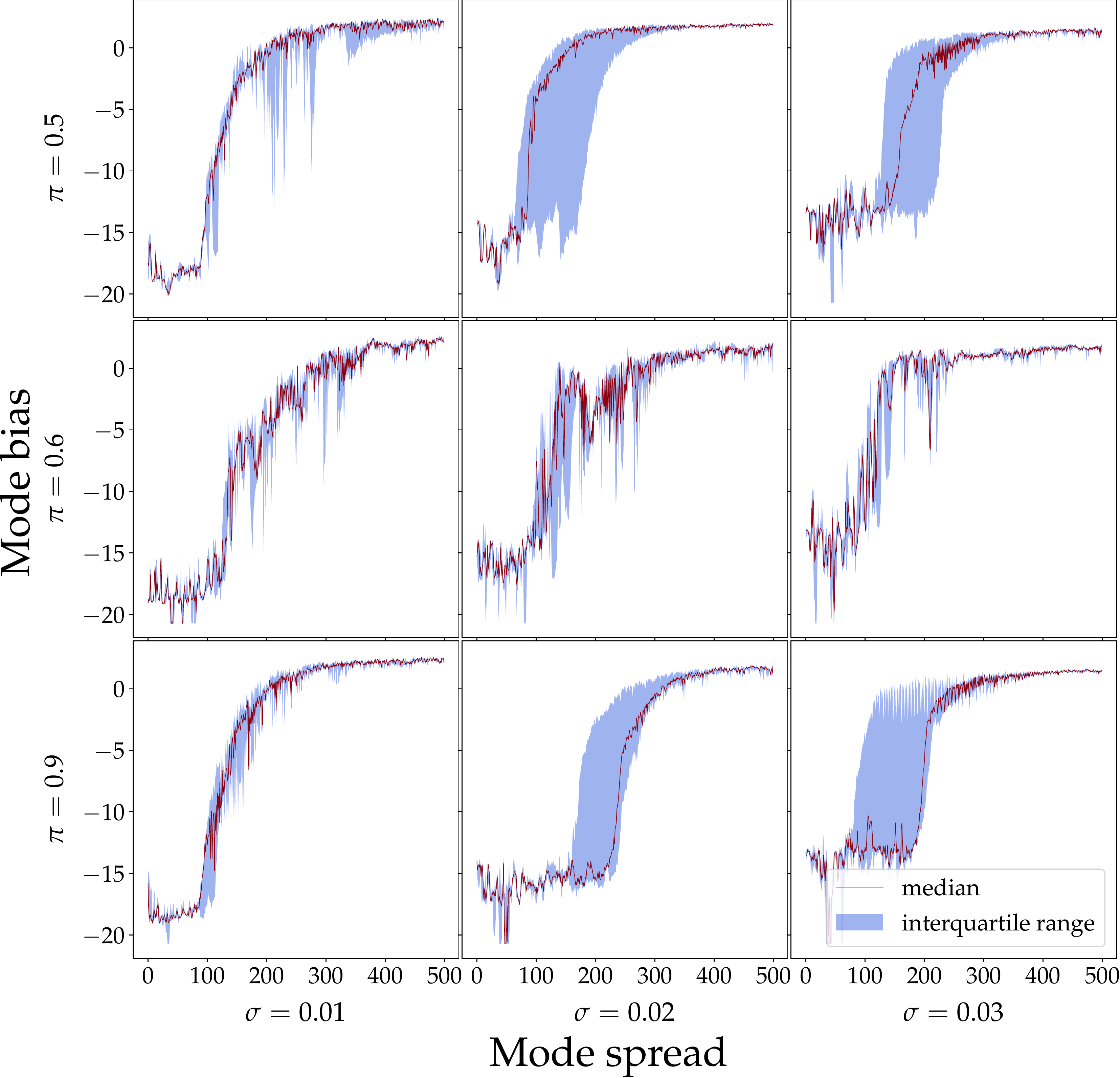}}
    \,
    \subfloat[]{\includegraphics[width=0.43\textwidth]{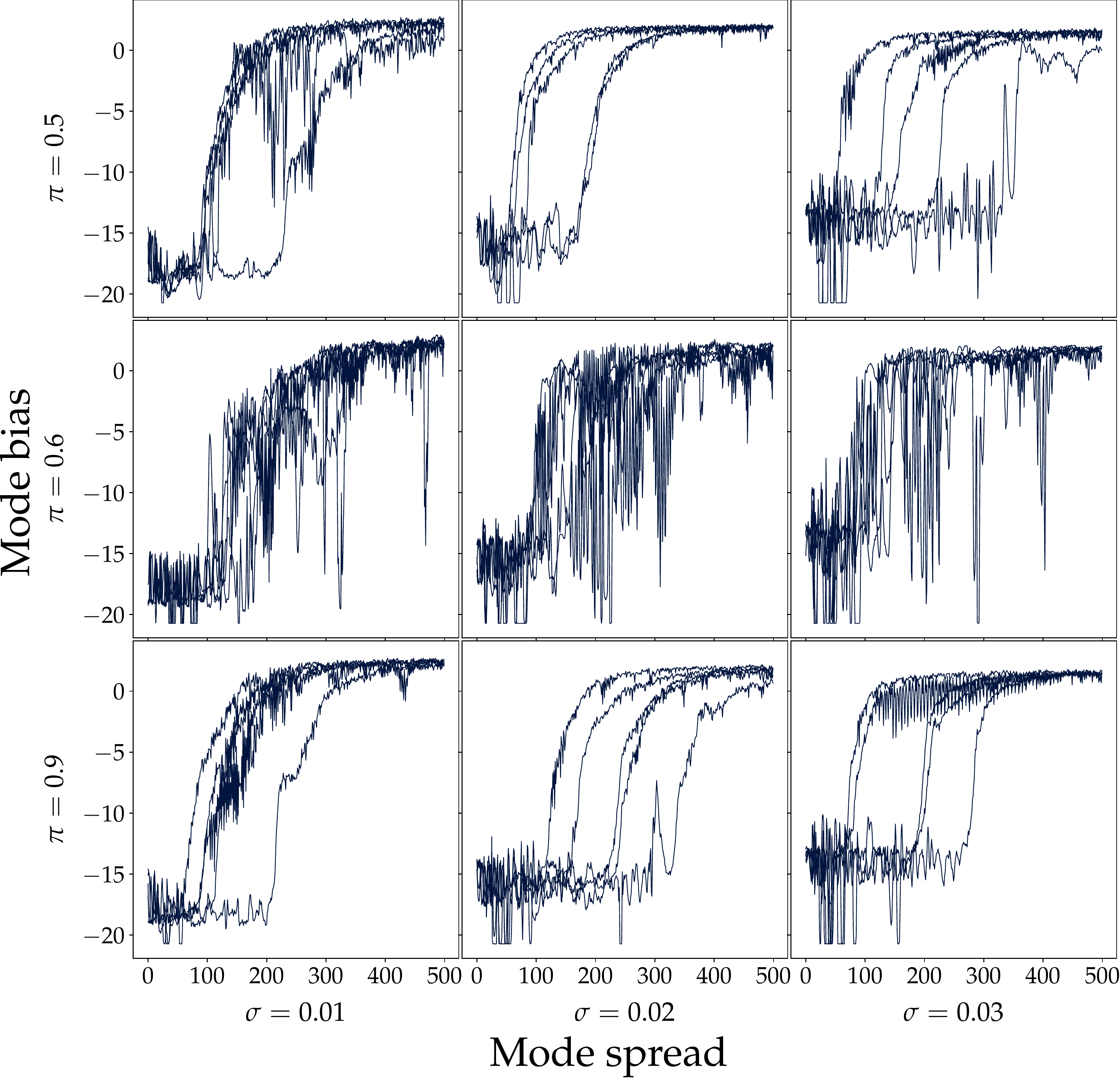}}
    \caption{Log-likelihood curves for 5 runs on each of the 9 data configurations. \textbf{(a)}~No calibration loss ($\lambda = 0$), averaged. \textbf{(b)}~No calibration loss, individual runs. \textbf{(c)}~With calibration loss ($\lambda = 1$), averaged. \textbf{(d)}~With calibration loss, individual runs.}%
    \label{fig:synth_configs}%
\end{figure}

\cref{fig:synth_configs} shows the data log-likelihoods for the 9 data configurations for varying mode bias $\pi \in \{0.5, 0.6, 0.9\}$ and mode noise $\sigma \in \{0.01, 0.02, 0.03\}$ trained with and without the calibration loss $\mathcal{L}_\text{cal}$. Each experiment is repeated 5 times and the individual likelihood curves are plotted in~\cref{fig:synth_configs}b and ~\cref{fig:synth_configs}d respectively. The results show that high bias is harder to learn, reflected by a slowed down convergence, however, the 
CAR model shows greater robustness to weight initialisation. In contrast the non-regularised GAN exhibits mode oscillation expressed as a fluctuation of higher likelihood (one mode is covered) and lower one (between modes). 

\subsection{LIDC}
\label[supplementary]{sup:more_lidc}
\subsubsection{Qualitative Analysis}

\begin{figure}[!b]
    \centering
    \includegraphics[width=0.80\textwidth]{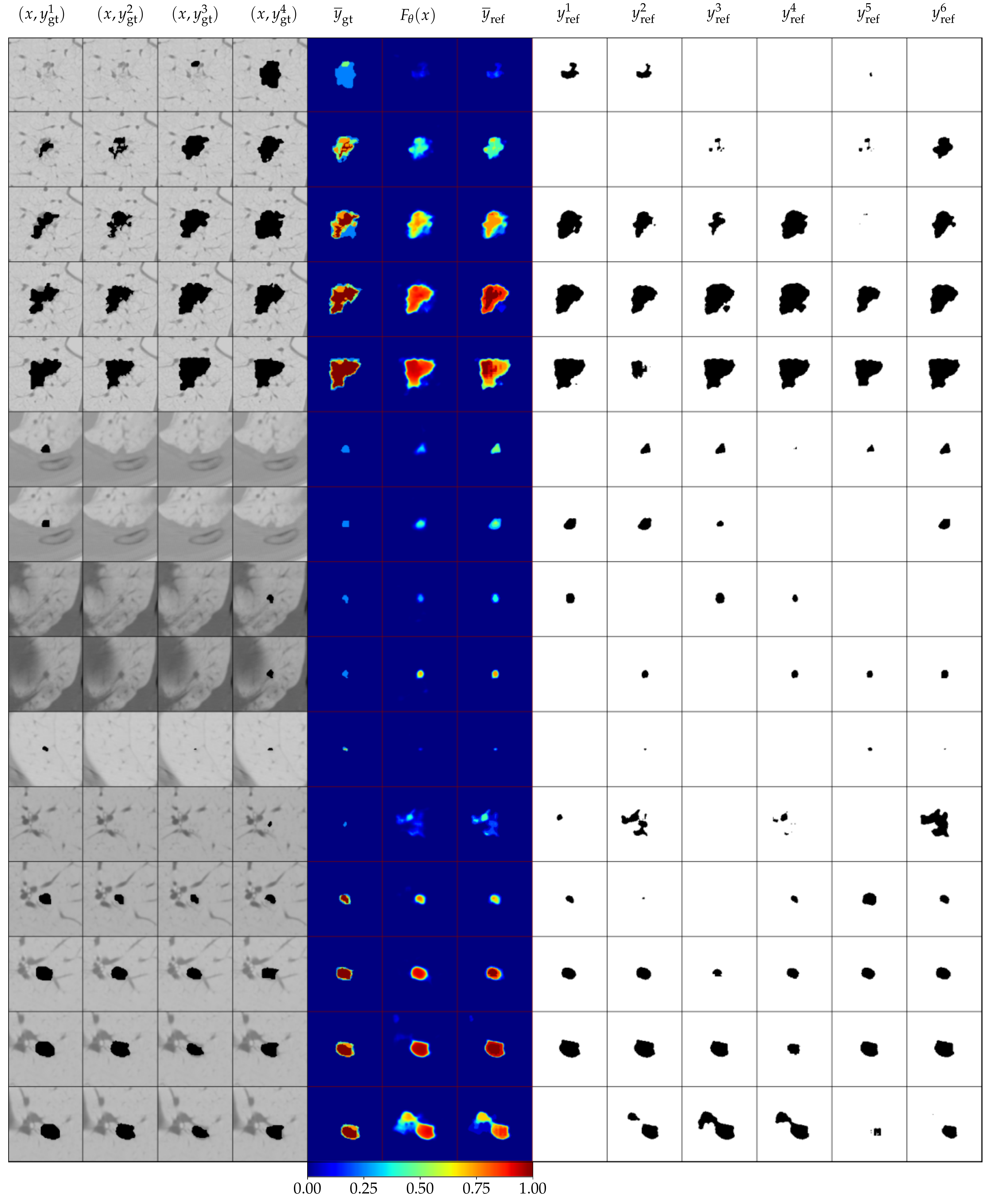}
    \caption{Qualitative results on LIDC samples for the CAR model.
    }%
    \label{fig:lidc_samples_1}%
\end{figure}

\begin{figure}[t!]
    \centering
    \includegraphics[width=0.8\textwidth]{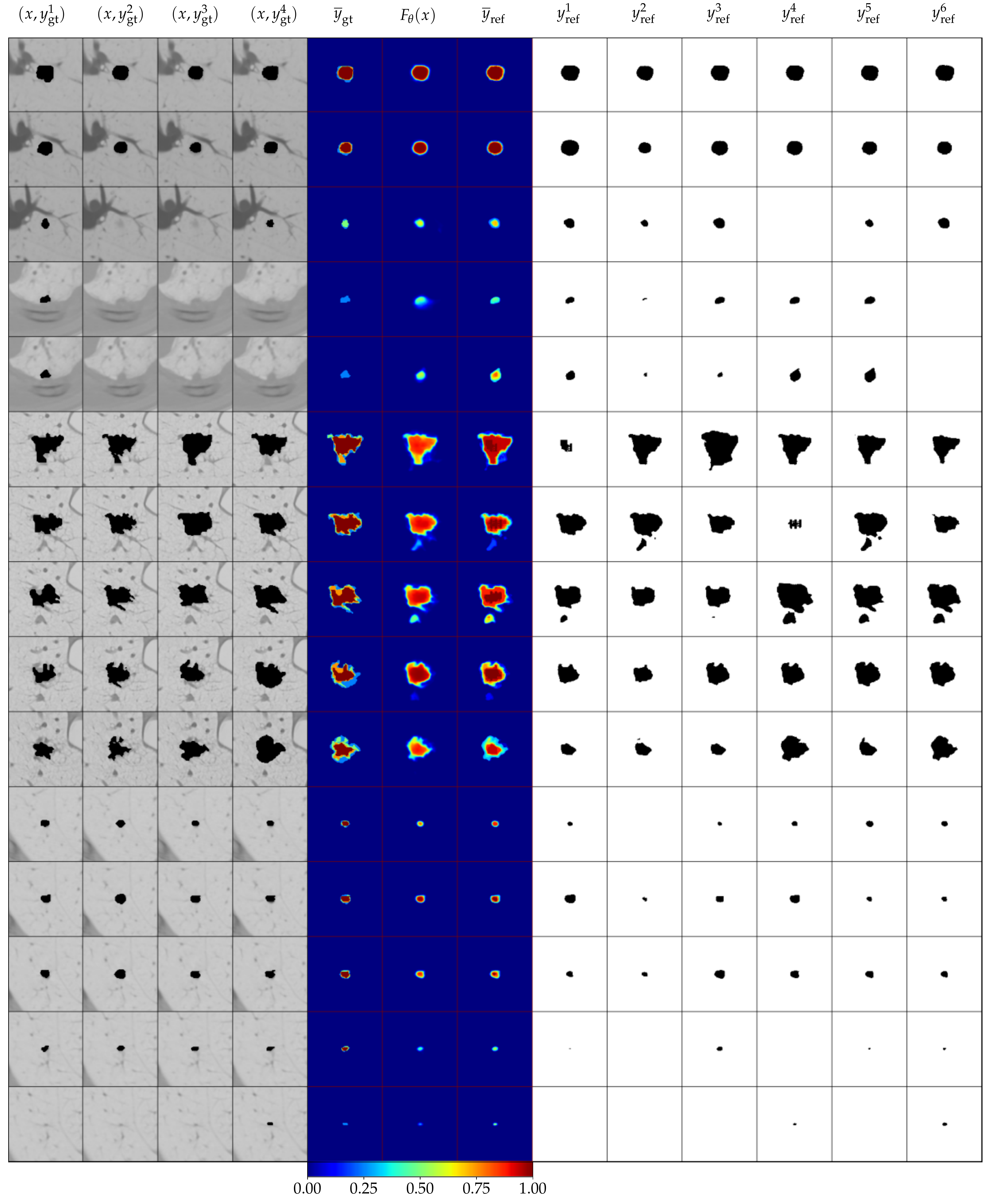}
    \caption{Qualitative results on LIDC samples for the CAR model.}
    \label{fig:lidc_samples_2}%
\end{figure}

To further examine the 
CAR model trained on the LIDC dataset, we illustrate representative qualitative results in~\cref{fig:lidc_samples_1} and~\cref{fig:lidc_samples_2}. For every input image $x$, we show the ground truth labels $y_{\text{gt}}^1,\ldots, y_{\text{gt}}^4$ provided by the different expert annotators, overlaying the input image, in the first four columns, and $6$ randomly sampled predictions $y_{\text{ref}}^1,\ldots,y_{\text{ref}}^6$ in the last six columns. From left to right, the three columns with the dark blue background in the center of the figures show the average ground truth predictions $\bar{y}_{\text{gt}}$, the output of the calibration network $F_\theta(x)$ and the average of 16 sampled predictions from the refinement network $\bar{y}_{\text{ref}}$. Our results show that even though there is a significant variability between the refinement network samples for a given input image, $\bar{y}_{\text{ref}}$ is almost identical to the calibration target $F_\theta(x)$, due to the diversity regularisation enforced by the calibration loss $\mathcal{L}_\mathrm{cal}$. 

From the qualitative results in~\cref{fig:lidc_samples_1} and~\cref{fig:lidc_samples_2}, it can be seen that the calibration target $F_\theta(x)$ does not always precisely capture the average of the ground truth distribution $\hat{y}_{\text{gt}}$, affecting the fidelity of the predictive distribution of the refinement network $G_\phi$. This indicates the importance of future work on improving the calibration of $F_\theta$, \eg implementing the approaches of~\cite{calibration,kull2019beyond}. 

\newpage

\subsubsection{Tuning the number of refinement network samples}

\begin{table}[H]
    \centering
      \begin{tabular}{lcccc}
        \toprule
        Method                 &  GED $\downarrow$ (16) & GED $\downarrow$ (50) & GED $\downarrow$ (100) & HM-IoU $\uparrow$ (16)\\ 
        \midrule
        cGAN+$\mathcal{L}_\mathrm{cal} \ (1)$ & $ 0.644 \pm 0.033$ & $ 0.643 \pm 0.033$ & $ 0.643 \pm 0.033 $ & $0.494 \pm 0.013$\\
        cGAN+$\mathcal{L}_\mathrm{cal} \ (5)$ & $0.278 \pm 0.000$ & $0.257 \pm 0.002$ & $0.252 \pm 0.001$ & $0.585 \pm 0.003$\\
        cGAN+$\mathcal{L}_\mathrm{cal} \ (10)$ & $ 0.277 \pm 0.003$ & $ 0.257 \pm 0.003$ & $ 0.250 \pm 0.003$ & $0.589 \pm 0.007$\\
        cGAN+$\mathcal{L}_\mathrm{cal} \ (15)$ & $ 0.271 \pm 0.002$ & $0.250 \pm 0.001$ & $0.245 \pm 0.003$ & $\mathbf{0.593 \pm 0.002}$\\
        cGAN+$\mathcal{L}_\mathrm{cal} \ (20)$ & $ \mathbf{0.264 \pm 0.002}$ & $\mathbf{0.248 \pm 0.004}$ & $\mathbf{0.243\pm 0.004}$ & $\mathbf{0.592 \pm 0.005}$\\
        \bottomrule
      \end{tabular}
      \caption{Mean GED and HM-IoU scores on LIDC for the CAR model ($\mathcal{L}_\text{cal}$-regularised cGAN)
       with 1, 5, 10, 15 and 20 samples. The number of samples used to compute the GED score is denoted in the parentheses in the header of each column. The arrows $\uparrow$ and $\downarrow$ denote if higher or lower score is better.}
    \label{tab:lidc_comaprison_n_samples}
\end{table}

To investigate the effect of the number of samples used to compute $\mathcal{L}_\mathrm{cal}$ on the learnt predictive distribution, we experimented on the CAR model using 5, 10, 15 or 20 samples from the refinement network $G_\phi$ during training. As a control experiment, we also train the same model using one sample. Our results, reported in~\cref{tab:lidc_comaprison_n_samples}, show that increasing the number of samples improves the quality of the predictive distribution, whereas using only one sample collapses it. This is expected because increasing the number of samples reduces the variance of the sample mean $\overline{G}_{\phi}$ and refines the approximation $\q[\phi]$ of the implicit predictive distribution realised by $G_\phi(x, \epsilon)$. 
Since in our implementation we reuse the samples from $G_\phi$ in the adversarial component $\mathcal{L}_{\mathrm{G}}$ of the total refinement network loss $\mathcal{L}_{\mathrm{G}}$, the discriminator $D_\psi$ interacts with a larger set of diverse fake samples during each training iteration, thus also improving the quality of $\mathcal{L}_{\mathrm{G}}$.

It is important to note that the benefit of increasing the sample size on the quality of $\mathcal{L}_\mathrm{cal}$ highly depends on the intrinsic multimodality in the data. In theory, if the number of samples used matches or exceeds the number of ground truth modes for a given input, it is sufficient to induce a calibrated predictive distribution. However, we usually do not have a priori access to this information. Conversely, if the sample size is too small, the $\mathcal{L}_\mathrm{cal}$ loss may introduce bias in the predictive distribution. This could lead to mode coupling or mode collapse, as exemplified in our control experiment with one sample.

In the LIDC dataset, even though we have access to four labels per input image, we argue that the dataset exhibits distributed multimodality, where a given pattern in the input space, \eg in a patch of pixels, can be associated to many different local labels throughout the dataset. As a result, an input image may correspond to more solutions than the four annotations provided. Therefore increasing the number of samples to more than four shows further improvement in performance. This however can come at the cost of decreased training speed which can be regulated by tuning the sample count parameter while considering the system requirements.

\subsubsection{Inducing multimodality in latent variable models on the LIDC dataset}
\label[supplementary]{sup:lidc_lvm_exps}

To examine whether conditioning the source of stochasticity in our model on the input is beneficial, we adapt our framework in order to learn a distribution over a latent code $z$, instead of taking samples from a fixed noise distribution $\epsilon \sim \gauss{0, 1}$, using variational inference~\cite{graves2011practical}. Following most of the existing work on stochastic semantic segmentation~\cite{rel:probUnet,kohl2019hierarchical,multipleannotations,baumgartner2019phiseg}, we maximise a modified lower bound on the conditional marginal log-likelihood, through a variational distribution $\q{z}{x, y}$. This is realised by minimising the loss function in~\cite{higgins2017beta}, given by:
\begin{equation}
    \mathcal{L}_{\text{ELBO}}(x, y) = -\expc[\q{z}{x, y}]{\log \p{y}{x, z}} + \beta\kl{\q{z}{x, y}}{\p{z}} \geq -\log \p{y}{x},
    \label{eq:b_vae}
\end{equation}
where $\beta$ controls the amount of regularisation from a prior distribution $\p{z}$ on the approximate posterior $\q{z}{x, y}$.
Both $\q{z}{x, y}$ and $\p{z}$ are commonly taken as factorised Gaussian distributions.

To this end, we compare our CAR model to two baselines where the refinement network $G_{\phi}$ is given as a cVAE-GAN~\cite{larsen2015autoencoding}. In the first one, we train $G_{\phi}$ by complementing the adversarial loss $\mathcal{L}_{\mathrm{adv}}$ with~\cref{eq:b_vae}, using $\beta \in \{0.1, 1, 10\}$ and a fixed standard normal prior. In the second, we introduce a calibration network and train $G_{\phi}$ using~\cref{eq:loss_total}. Note that this does not necessitate specifying a prior.

For a fair comparison, we use the same core models for all of our experiments, introducing only minor modifications to the refinement network to convert the deterministic encoder into a probabilistic one with Gaussian output. This is achieved by splitting the output head of the encoder so as to predict the mean and standard deviation of the encoded distribution~\cite{vae, kendall2017uncertainties}. Instead of using random noise sampled from a standard Gaussian as our source of stochasticity, the decoder of the refinement network is now injected with latent codes sampled from the Gaussian distribution encoded for each input image. To train the model, we pretrain the $F_{\theta}$ in isolation, and subsequently apply it in inference mode while training $G_{\phi}$. We use a batch size of 32, an 8-dimensional latent code, and use 20 samples to compute $\mathcal{L}_\mathrm{cal}$.

\begin{table}[t!]
    \centering
    \begin{threeparttable}
    \addtolength{\tabcolsep}{-1pt}
      \begin{tabular}{lcccc}
        \toprule
        Method                 &  GED $\downarrow$ (16) & GED $\downarrow$ (50) & GED $\downarrow$ (100) & HM-IoU $\uparrow$ (16) \\ 
        \midrule
        cGAN+$\mathcal{L}_\mathrm{ce}$ & $0.639 \pm 0.002$  & \textemdash & \textemdash & $0.477 \pm 0.004$ \\
        CAR (ours) & $\mathbf{0.264 \pm 0.002}$ & $\mathbf{0.248 \pm 0.004}$ & $0.243\pm 0.004$ & $\mathbf{0.592 \pm 0.005}$\\
        \midrule
        cVAE-GAN ($\beta$=$0.1$) & $0.577 \pm 0.095$ & \textemdash & \textemdash & $0.484 \pm 0.006$\\
        cVAE-GAN ($\beta$=$1$) & $0.596 \pm 0.078$ & \textemdash & \textemdash & $0.474 \pm 0.005$\\
        cVAE-GAN ($\beta$=$10$) & $0.609 \pm 0.061$ & \textemdash & \textemdash & $0.482 \pm 0.010$\\
        cVAE-GAN+$\mathcal{L}_\mathrm{cal}$ ($\beta$=$0$) & $0.272 \pm 0.006$ & $0.252 \pm 0.006$ & $0.246 \pm 0.006$ & $\mathbf{0.593 \pm 0.003}$\\
        \bottomrule
      \end{tabular}
    \end{threeparttable}
    \caption{GED and HM-IoU scores on LIDC. The top section shows the $\mathcal{L}_\text{ce}$-regularised baseline and the CAR model; the bottom section shows baseline and $\mathcal{L}_\text{cal}$-regularised cVAE-GANs. All $\mathcal{L}_\text{cal}$-regularised models are trained using $20$ samples. The three central columns show the GED score computed with 16, 50 and 100 samples, respectively. The last column shows the HM-IoU score, computed with 16 samples. 
    The arrows $\uparrow$ and $\downarrow$ indicate whether higher or lower score is better.}
    \label{tab:lidc_comparison2}
\end{table}

\begin{figure}[h]%
    \centering
    \subfloat[]{\includegraphics[width=\textwidth]{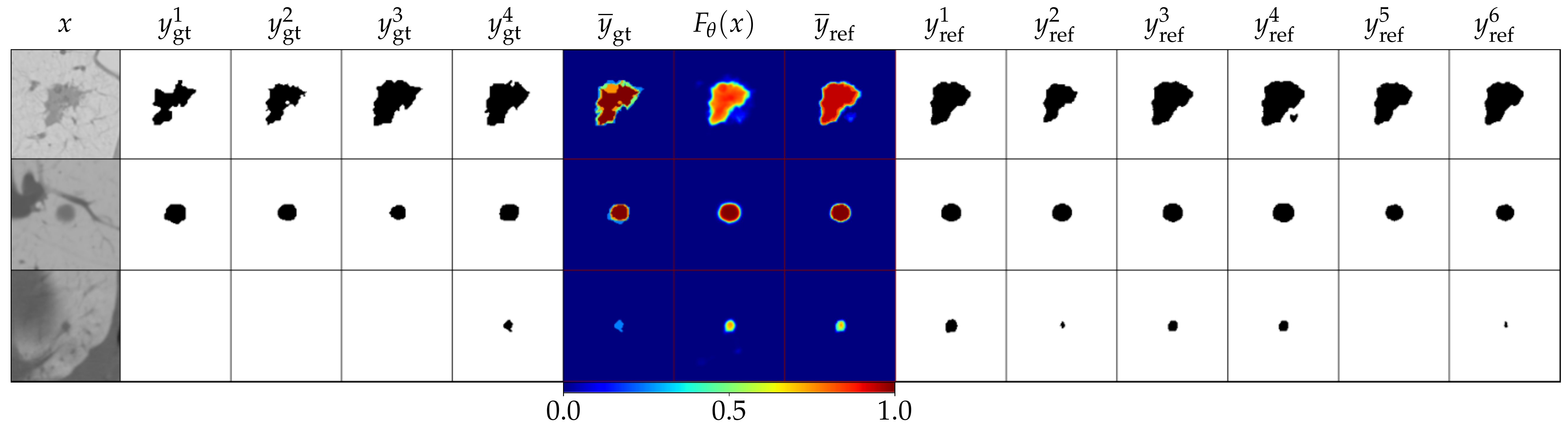}}
    \\
    \subfloat[]{\includegraphics[width=\textwidth]{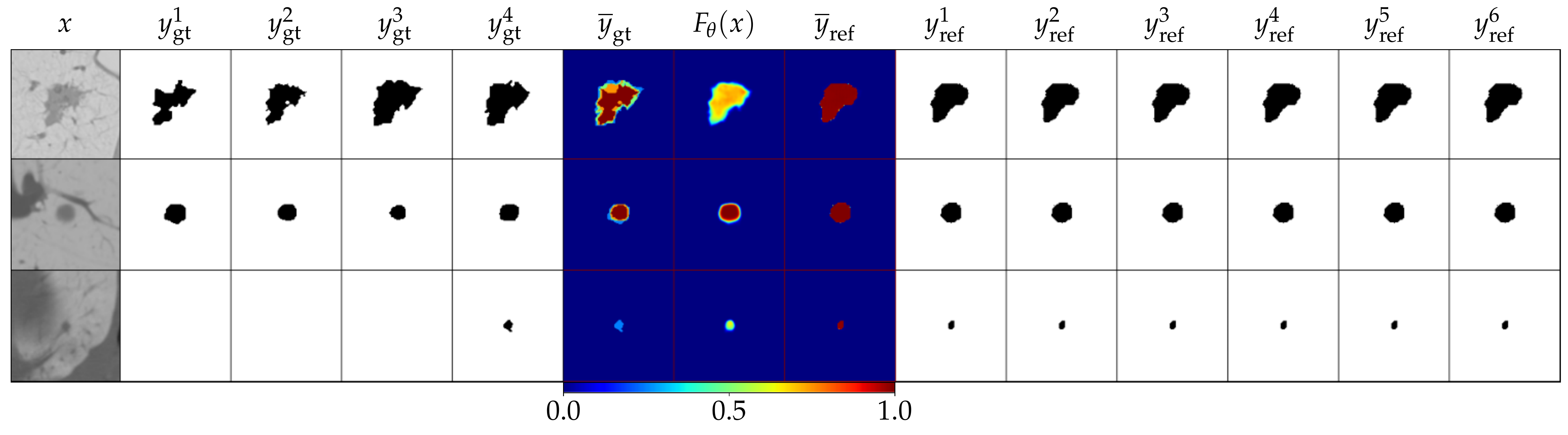}}
    \caption{LIDC validation samples for the \textbf{(a)} cVAE-GAN and \textbf{(b)} cGAN+$\mathcal{L}_\text{ce}$ baseline model.
    }%
    \label{fig:lidc_samples_vae}%
\end{figure}

We show that the cVAE-GAN model trained with $\mathcal{L}_\mathrm{cal}$ instead of the traditional complexity loss term, $\kl{\q{z}{x, y}}{\p{z}}$ from~\cref{eq:b_vae} is able to learn a distribution over segmentation maps, and performs similarly to the CAR model. This is important because it abrogates the need for specifying a latent-space prior, which is often selected for computational convenience, rather than task relevance~\cite{NCPs, FNP}. On the other hand, our cVAE-GAN models trained using the KL-divergence complexity term showed limited diversity, even for large $\beta$ values. The results, shown quantitatively in the bottom part of~\cref{tab:lidc_comparison2}, 
and qualitatively in~\cref{fig:lidc_samples_vae}a,
demonstrate that the interaction between $\mathcal{L}_{\mathrm{adv}}$ and $\mathcal{L}_\mathrm{cal}$ can sufficiently induce a multimodal predictive distribution in latent variable models, and indicate that for the purpose of stochastic semantic segmentation the use of a probabilistic encoder is not strictly required. 

\subsection{Cityscapes}
\label[supplementary]{sup:more_cs}
\begin{figure}[b!]
    \centering
    \subfloat{\includegraphics[height=3.65cm]{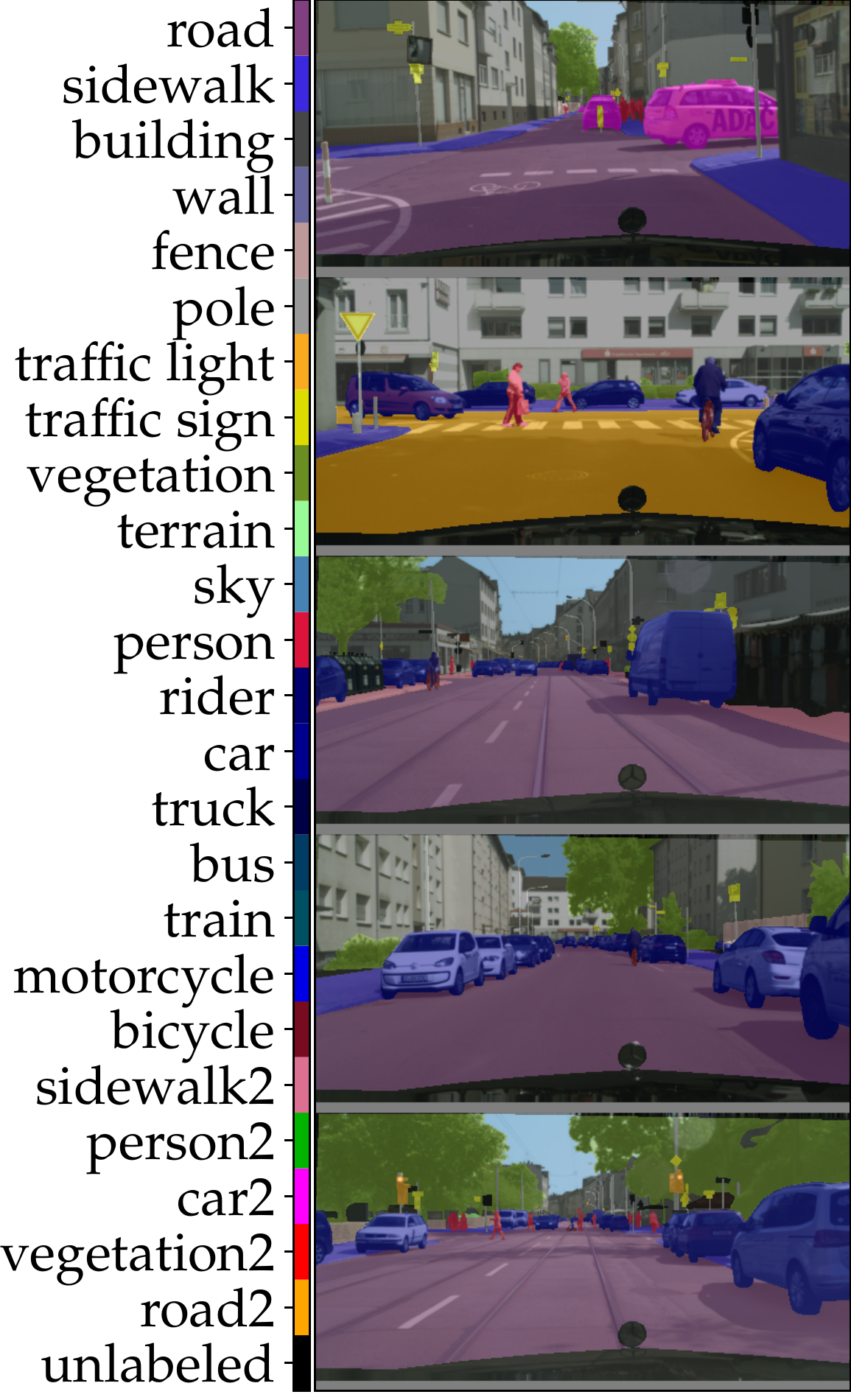}}
    \subfloat{\includegraphics[height=3.65cm]{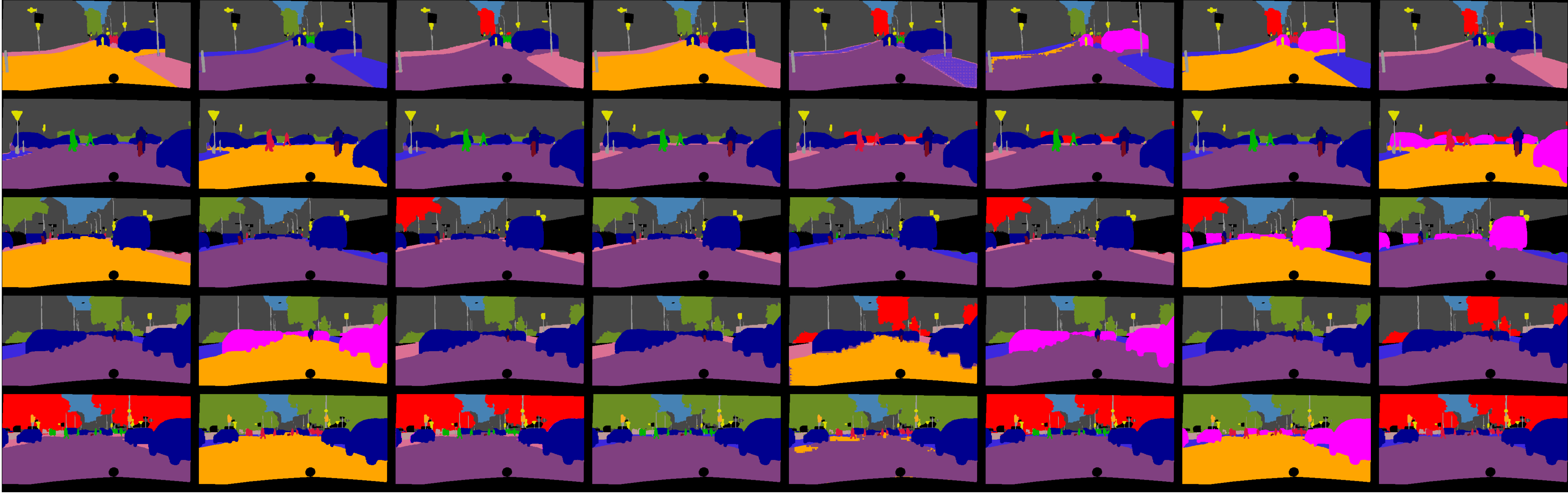}}
    \\
    \hspace{0.33cm}\subfloat{\includegraphics[height=3.65cm]{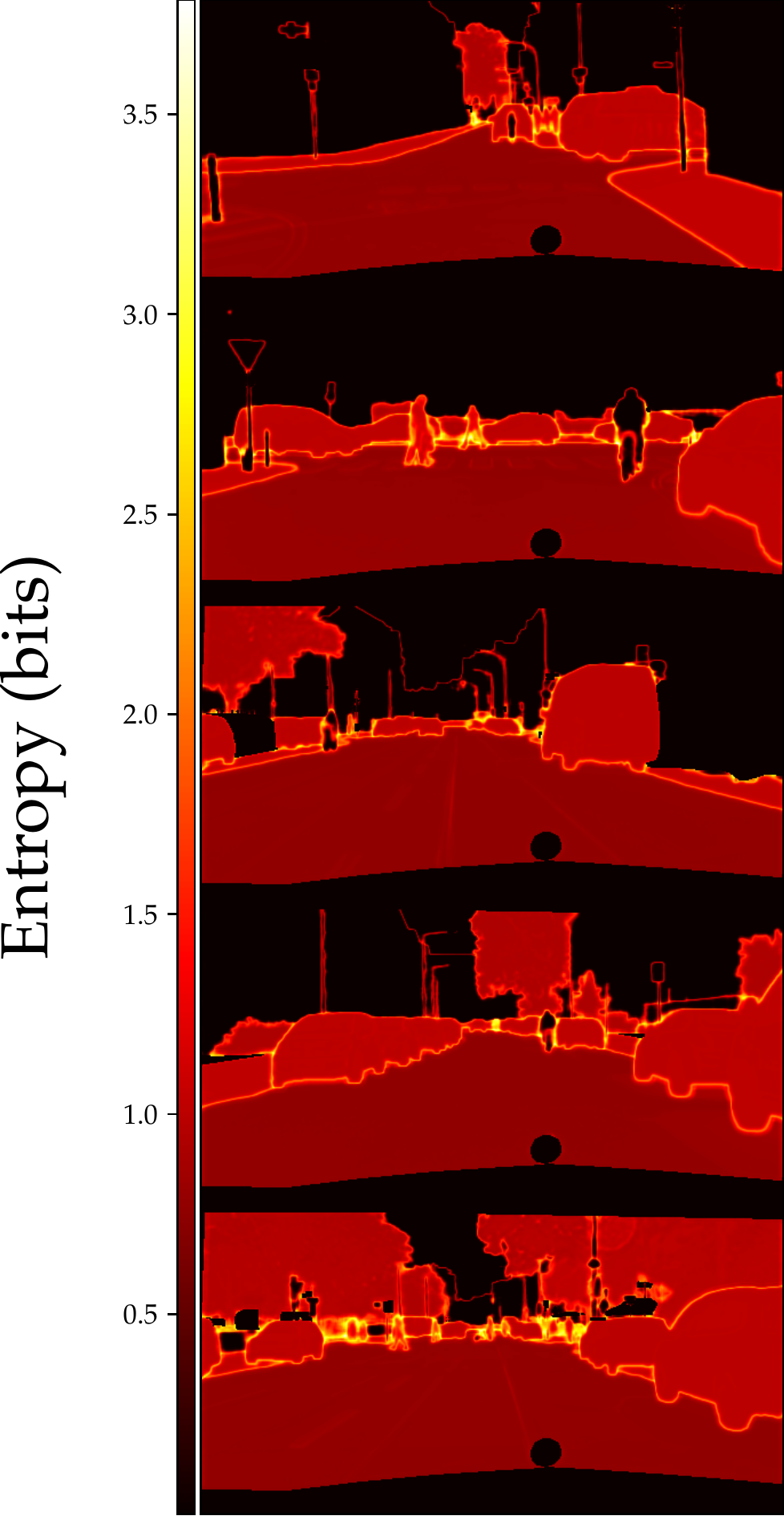}}
    \subfloat{\includegraphics[height=3.65cm]{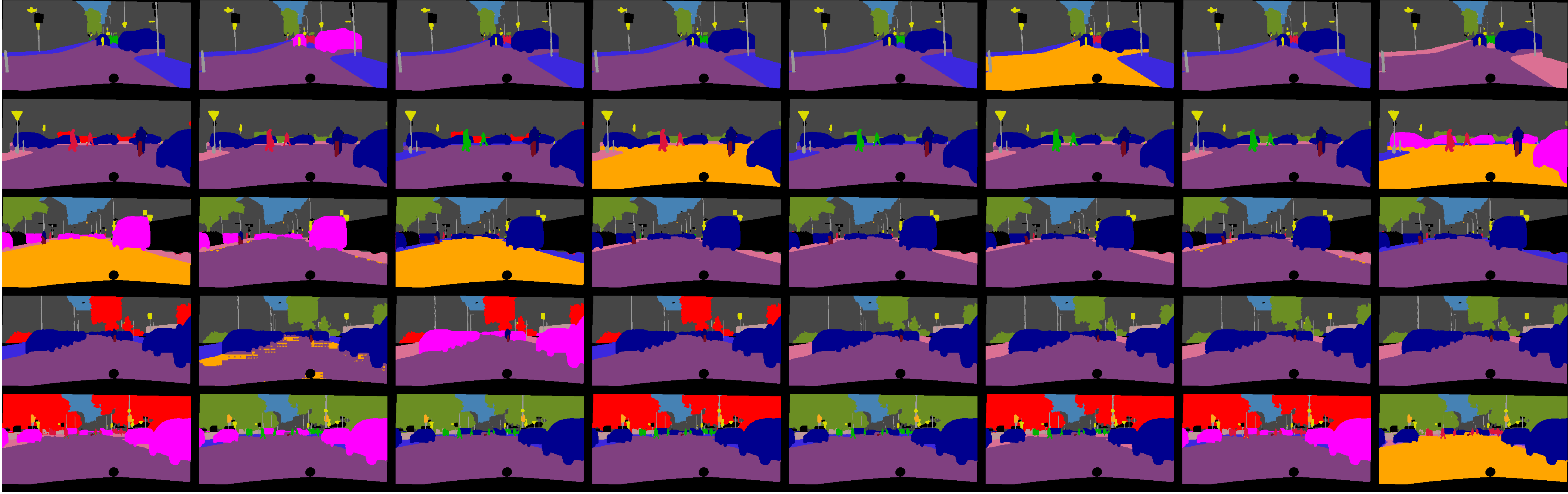}}
    \\
    \vspace{1cm} 
    \subfloat{\includegraphics[height=3.65cm]{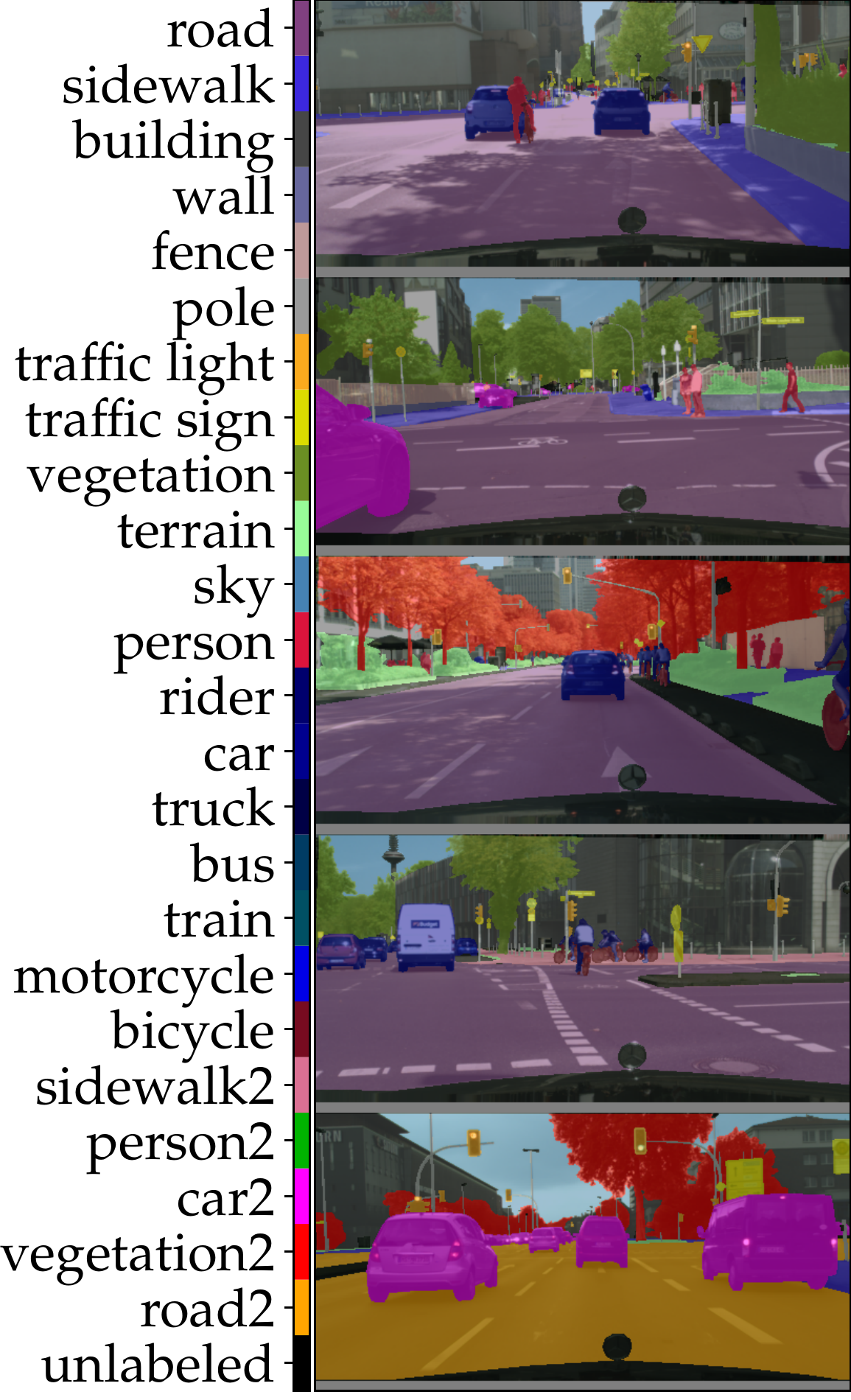}}
    \subfloat{\includegraphics[height=3.65cm]{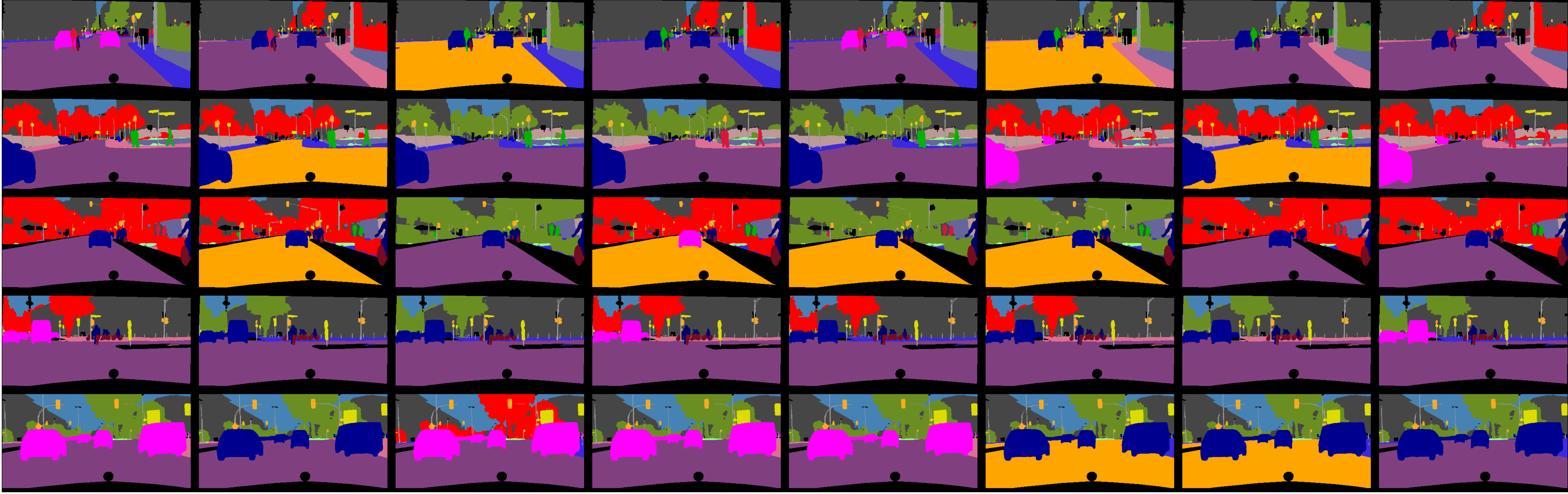}}
    \\
    \hspace{0.33cm}\subfloat{\includegraphics[height=3.65cm]{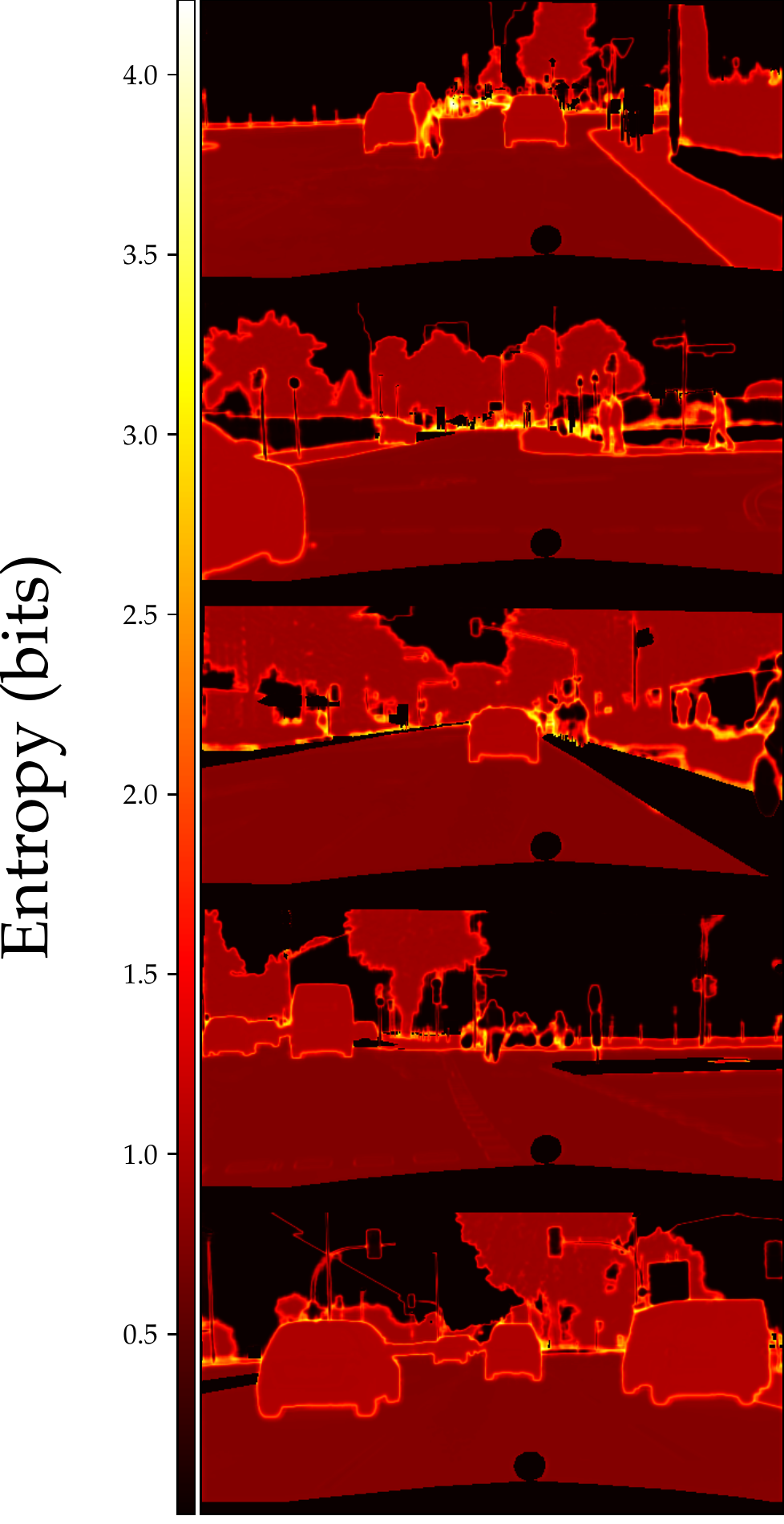}}
    \subfloat{\includegraphics[height=3.65cm]{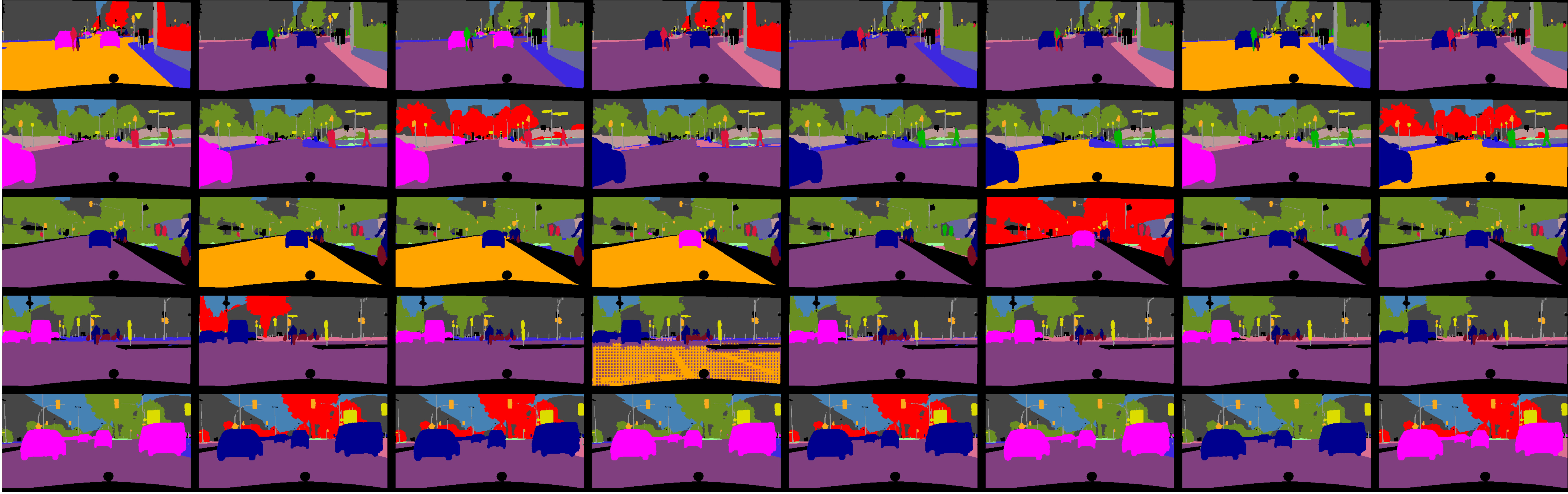}}
    \caption{10 input images, the corresponding aleatoric maps from the calibration network and 16 samples from the refinement network. For visualisation purposes, the samples are split into 8 per row.}%
    \label{fig:cs_samples}%
\end{figure}

\subsubsection{Qualitative Analysis}

In this section we provide additional qualitative results for the CAR model trained on the modified Cityscapes dataset~\cite{rel:probUnet}. In~\cref{fig:cs_samples}, we show 16 randomly sampled predictions for representative input images $x$, and their corresponding aleatoric uncertainty maps, obtained by computing the entropy of the output of the calibration network, $\entropy{F_\theta(x)}$, as done in~\cite{kendall2017uncertainties}.
The predicted samples are of high quality, evident by object coherence and crisp outlines, and high diversity, where all classes are well represented. Our model effectively learns the entropy of the ground truth distribution in the stochastic classes (\textit{sidewalk}, \textit{person}, \textit{car}, \textit{vegetation} and \textit{road}), as their distinct entropy levels are captured as different shades of red in the entropy maps, corresponding to the different flip probabilities ($\nicefrac{8}{17}$, $\nicefrac{7}{17}$, $\nicefrac{6}{17}$, $\nicefrac{5}{17}$ and $\nicefrac{4}{17}$ respectively). Additionally, it can be seen that edges or object boundaries are also highlighted in the aleatoric uncertainty maps, reflecting inconsistency during manual annotation, which often occurs on input pixels that are difficult to segment.

\begin{figure}[b!]
    \centering
    \hspace{0.3cm}
    \subfloat[]{\includegraphics[height=2.2cm]{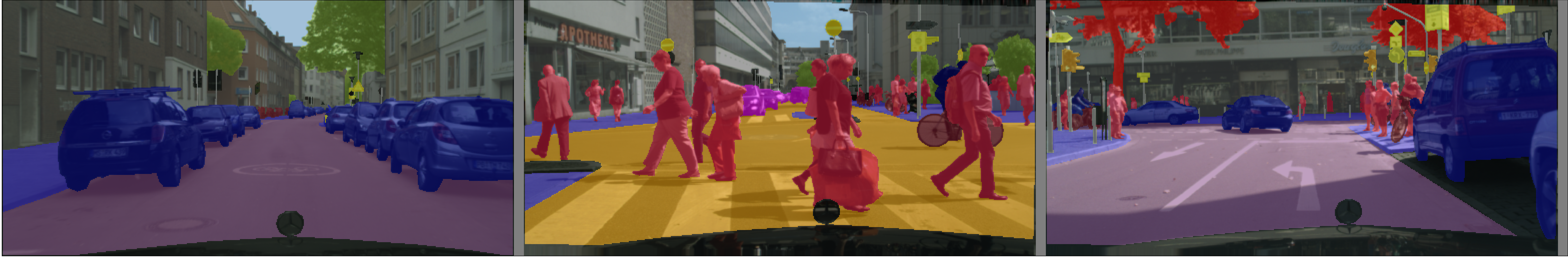}}
    \\
    \hspace{0.3cm}
    \subfloat[]{\includegraphics[height=2.2cm]{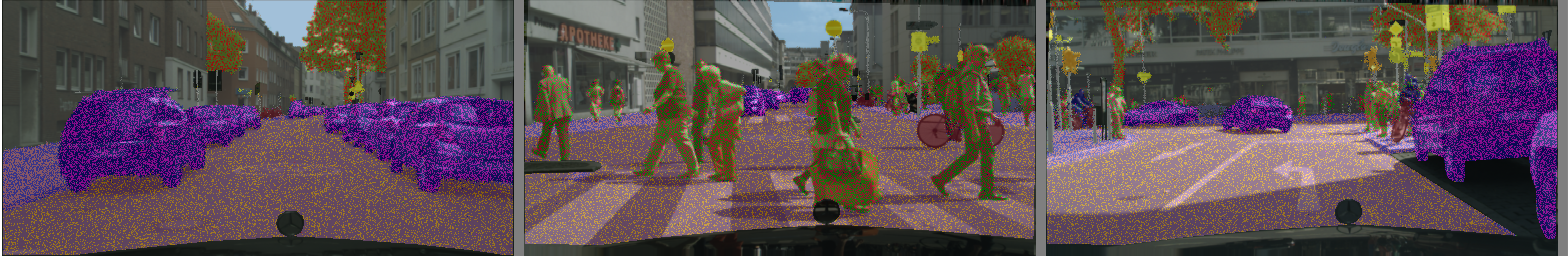}}
    \\
    \subfloat[]{\includegraphics[height=2.2cm]{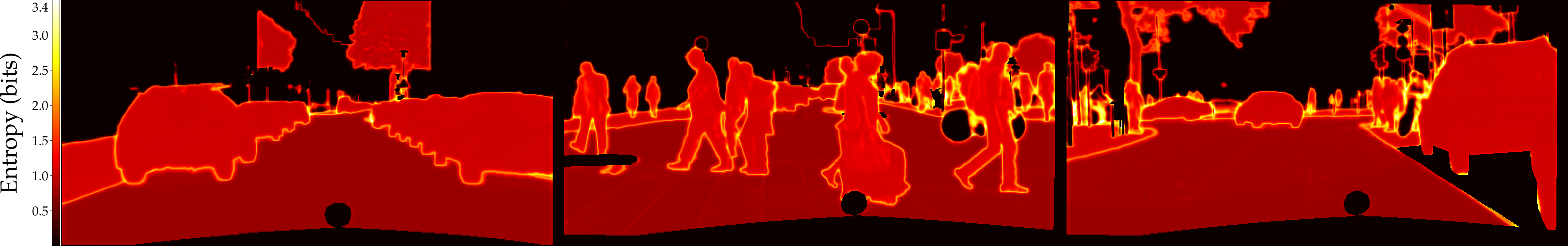}}
    \\
    \subfloat[]{\includegraphics[height=2.2cm]{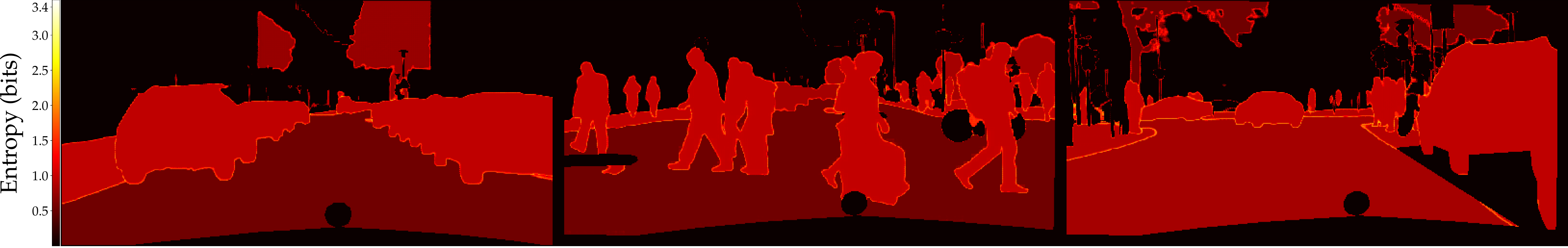}}
    \\
    \subfloat[]{\includegraphics[height=2.2cm]{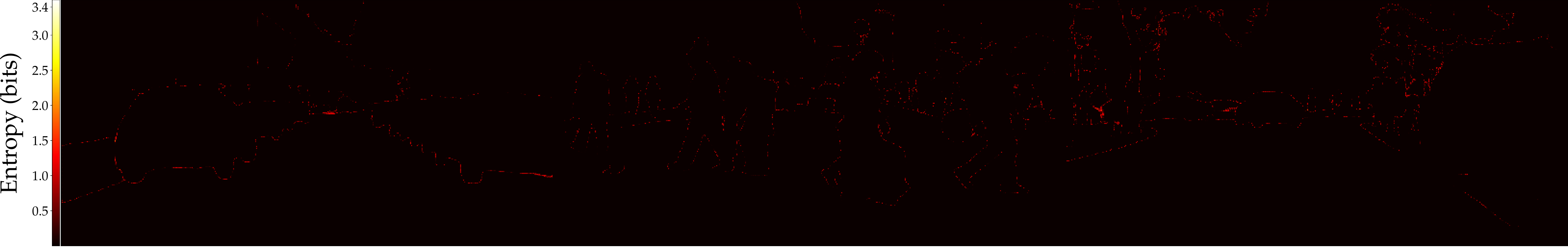}}
    \caption{
    \textbf{(a)} Three input images overlaid with the corresponding labels; 
    \textbf{(b)} Incoherent samples from the predictive distribution of the calibration network;
    \textbf{(c)} The aleatoric maps from the calibration network; 
    \textbf{(d)} Aleatoric maps computed as the entropy of the average of 16 predictions of the refinement network; 
    \textbf{(e)} The entropy of one sample of the refinement network output for each input image.}%
    \label{fig:cs_entropy}%
\end{figure}

\cref{fig:cs_entropy}d shows the entropy of the predictive distribution of the refinement network $G_\phi$, $\entropy{\overline{G}_{\phi}(F_\theta(x))}$, where $\overline{G}_{\phi}(F_\theta(x))$ is computed as the average of 16 samples from $G_\phi$. Our results demonstrate that $\entropy{\overline{G}_{\phi}(F_\theta(x))}$ is similar to $\entropy{F_\theta(x)}$, depicted in \cref{fig:cs_entropy}c, as encouraged by the $\mathcal{L}_\mathrm{cal}$ regularisation. Notice that object boundaries are also highlighted in $\entropy{\overline{G}_{\phi}(F_\theta(x))}$, indicating that our model captures shape ambiguity as well as class ambiguity. However, some uncertainty information from $\entropy{F_\theta(x)}$ is not present in $\entropy{\overline{G}_{\phi}(F_\theta(x))}$, \eg the entropy of the different stochastic classes are not always consistent across images, as evident from the different shades of red seen for the road class in~\cref{fig:cs_entropy}d. We expect that increasing the number of samples from the refinement network will improve the aleatoric uncertainty estimates. Nevertheless, the sample-free estimate extracted from $F_\theta(x)$ is cheaper to obtain and more reliable than the sample-based average from $G_\phi$, highlighting an important benefit of our cascaded approach.

Finally, we illustrate in~\cref{fig:cs_entropy}e the high confidence of the predictions from the refinement network $G_\phi(F_\theta(x))$, reflected by their low entropy, $\entropy{G_\phi(F_\theta(x))}$. This is attributed to the adversarial component in the refinement loss function, which encourages confident predictions to mimic the one-hot representation of the ground truth annotations. Even though each prediction of the refinement network is highly confident, the average of the predictions $\overline{G}_{\phi}(F_\theta(x))$ is calibrated, as shown in~\cref{fig:cs_entropy}d. This is a clear illustration of the advantage of complementing the adversarial loss term $\mathcal{L}_{\mathrm{G}}$ with the calibration loss term $\mathcal{L}_{\mathrm{cal}}$ in the training objective for the refinement network.

\end{appendices}

\end{document}